\documentclass[10pt,twocolumn,letterpaper]{article}

\usepackage[pagenumbers]{cvpr} 

\newcommand{\TODO}[1]{}

\usepackage{microtype}

\usepackage{amsmath, amssymb}

\usepackage{pifont}
\usepackage{booktabs}
\usepackage{multirow}
\usepackage[table]{xcolor}
\usepackage{caption}
\usepackage{subcaption}
\usepackage{graphicx}
\usepackage{tabularx}
\newcolumntype{Y}{>{\centering\arraybackslash}X}
\usepackage{array}
\usepackage{enumitem}

\definecolor{cvprblue}{rgb}{0.21,0.49,0.74}
\usepackage[pagebackref,breaklinks,colorlinks,allcolors=cvprblue]{hyperref}

\def\paperID{32845} 
\def\confName{CVPR}
\def\confYear{2026}

\title{MUST: Modality-Specific Representation-Aware Transformer for Diffusion-Enhanced Survival Prediction with Missing Modality}

\author{
Kyungwon Kim$^{1}$ \quad\quad
Dosik Hwang$^{1,2}$\thanks{Corresponding author} \\
$^{1}$ Yonsei University \quad
$^{2}$ Korea Institute of Science and Technology \\
{\tt\small \{yskgw93, dosik.hwang\}@yonsei.ac.kr}
}

\begin{document}
\maketitle
\begin{abstract}
Accurate survival prediction from multimodal medical data is essential for precision oncology, yet clinical deployment faces a persistent challenge: modalities are frequently incomplete due to cost constraints, technical limitations, or retrospective data availability. While recent methods attempt to address missing modalities through feature alignment or joint distribution learning, they fundamentally lack explicit modeling of the unique contributions of each modality as opposed to the information derivable from other modalities. We propose MUST (Modality-Specific representation-aware Transformer), a novel framework that explicitly decomposes each modality's representation into modality-specific and cross-modal contextualized components through algebraic constraints in a learned low-rank shared subspace. This decomposition enables precise identification of what information is lost when a modality is absent. For the truly modality-specific information that cannot be inferred from available modalities, we employ conditional latent diffusion models to generate high-quality representations conditioned on recovered shared information and learned structural priors. Extensive experiments on five TCGA cancer datasets demonstrate that MUST achieves state-of-the-art performance with complete data while maintaining robust predictions in both missing pathology and missing genomics conditions, with clinically acceptable inference latency. Our project page is available at \url{https://kylekwkim.github.io/MUST/}.
\end{abstract}
    
\section{Introduction}

Leveraging multiple data modalities has been shown to substantially improve performance across various domains~\cite{baltrusaitis2018multimodal,biomedclip,imagebind}. In survival prediction, integrating diverse information sources—such as whole slide images (WSI), genomic profiles, and other clinical data—enables more comprehensive prognostic models than single-modality approaches~\cite{chen2022pan,vale2021long,cheerla2019deep}. Among these, WSI capturing tissue morphology and genomic data revealing molecular alterations have emerged as particularly critical~\cite{chen2020pathomic}, as improved survival prediction directly translates to better risk stratification for treatment planning~\cite{acosta2022multimodal}. However, a critical gap exists between research settings with complete multimodal data and clinical reality where modalities are frequently incomplete.

In practice, genomic profiling remains expensive and time-consuming, limiting its routine clinical use. Historical patient cohorts often contain extensive histopathology archives without corresponding molecular data. Conversely, emerging precision medicine initiatives may have genomic data without complete tissue samples. This incompleteness poses a fundamental challenge: existing multimodal models typically assume complete data availability and exhibit severe performance degradation with missing modalities~\cite{acosta2022multimodal,li2022hierarchical}.

Current approaches to handling missing modalities can be categorized into three paradigms. \textit{Feature alignment methods}~\cite{wang2023mcf,radford2021learning,smil} train models to produce similar representations regardless of which modalities are available, but do not explicitly identify what information is missing—they merely enforce output similarity without modeling the semantic structure of cross-modal relationships. \textit{Imputation-based methods}~\cite{dorent2019hetero,sharma2019missing} attempt to synthesize missing features through generative models, but often operate in high-dimensional feature spaces, producing noisy pseudo-features that can degrade downstream performance. \textit{Joint distribution learning methods} like LD-CVAE~\cite{ldcvae} learn joint posteriors for conditional generation, capturing modality interactions but failing to disentangle what each modality uniquely contributes versus what can be derived from others. While information disentanglement has been studied in multimodal learning~\cite{daunhawer2023identifiability,lee2021private}, these approaches focus on representation interpretability rather than enabling algebraic recoverability of missing components. A fundamental limitation shared by existing methods is the absence of explicit modeling of what information is modality-specific versus cross-modally inferable.

We propose MUST (Modality-Specific representation-aware Transformer), which explicitly decomposes each modality's global representation into modality-specific and cross-modal contextualized components through bidirectional cross-attention and algebraic constraints in a learned low-rank shared subspace. Unlike unconstrained cross-attention methods (e.g., SurvPath~\cite{survpath}, CMTA~\cite{cmta}) that lack missing modality mechanisms, and unlike disentanglement approaches that separate components for interpretability alone, our framework enforces algebraic invertibility: shared components are deterministically derivable from any available modality. For truly modality-specific information, we train conditional latent diffusion models~\cite{ddpm,ldm} to generate plausible representations conditioned on recovered shared information and structural priors~\cite{syndiff,chung2022diffusion}. The main contributions are:

$\bullet$ An explicit decomposition architecture with bidirectional cross-attention and algebraic constraints in a low-rank shared subspace that identifies precisely what information is missing when a modality is absent, enabling deterministic recovery of shared components.

$\bullet$ A two-stage reconstruction strategy combining deterministic algebraic projection for shared information with LDM-based probabilistic generation for modality-specific information, isolating stochastic generation to truly modality-specific residuals only.

\noindent Extensive experiments on five TCGA cancer datasets demonstrate that MUST achieves state-of-the-art performance with complete data (C-index: 0.742) while maintaining robust predictions under missing genomics (0.716) and missing pathology (0.739), substantially outperforming existing methods in all scenarios.

\begin{figure*}[t]
  \centering
   \includegraphics[width=1.0\linewidth]{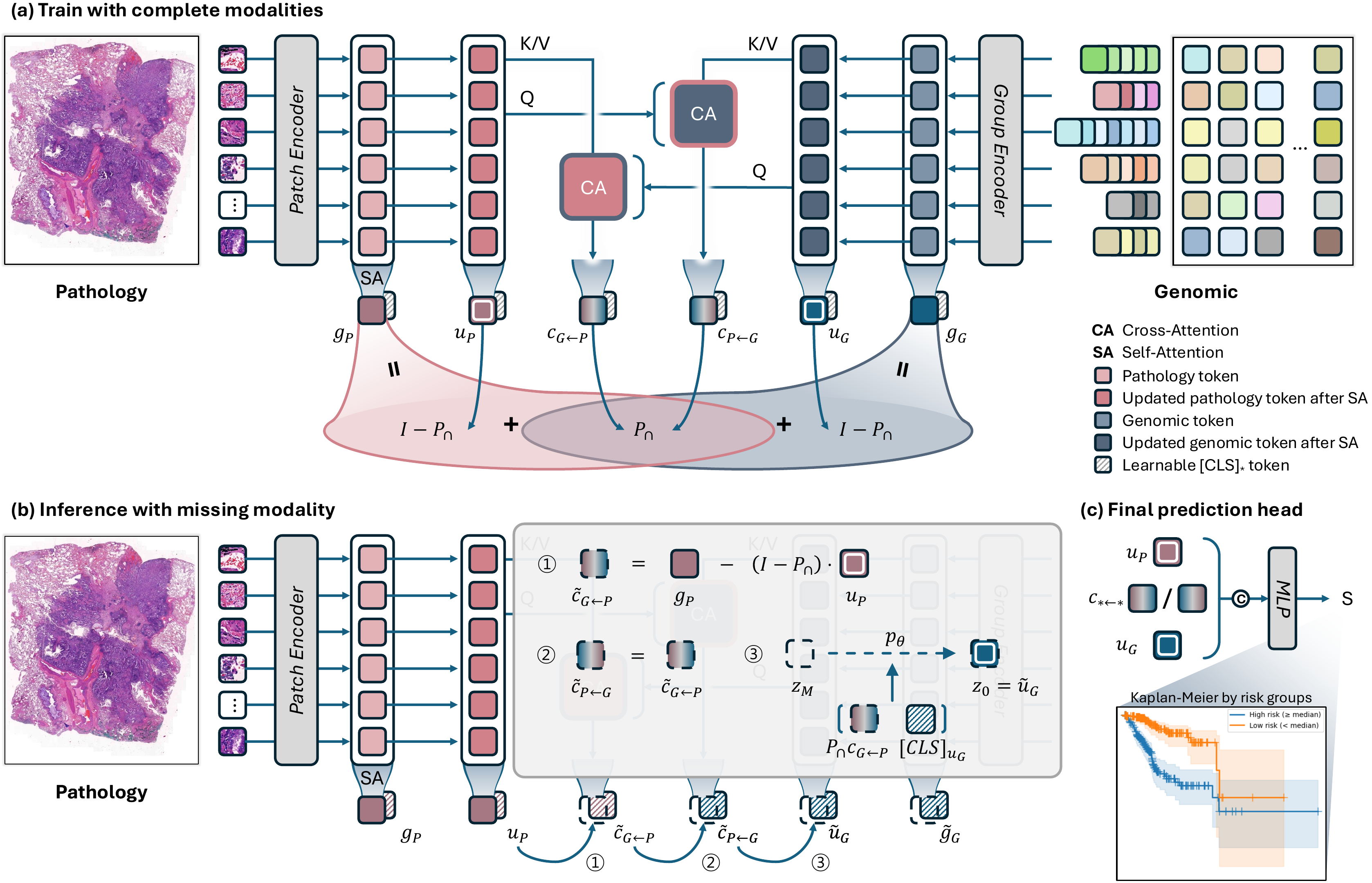}
   \caption{Overall architecture of MUST. The framework extracts global representations $g_P$ and $g_G$ via self-attention, computes shared information through bidirectional cross-attention, and decomposes each modality into specific and shared components for survival prediction.}
   \label{fig:framework}
\end{figure*}

\section{Related Work}

\subsection{Multimodal Survival Prediction}

Deep learning has revolutionized survival prediction in computational pathology, from CNNs for automatic feature extraction from gigapixel WSIs~\cite{campanella2019clinical} to attention-based methods like CLAM~\cite{clam} and TransMIL~\cite{transmil} for patch-level aggregation.

For multimodal integration, Chen et al.~\cite{chen2020pathomic} systematically compared early, late, and intermediate fusion across cancer types. Chen et al.~\cite{mcat} employ multimodal co-attention transformers to capture fine-grained interactions between pathology and genomic features. Jaume et al.~\cite{survpath} leverage foundation models for enhanced feature extraction, while Zhang et al.~\cite{cmta} use cross-modal transformers for attention-based fusion. Lipkova et al.~\cite{motcat} proposed multi-omic transformers for integrating diverse molecular data. However, these methods assume complete modality availability and cannot handle missing data scenarios, limiting their clinical applicability.

\subsection{Handling Missing Modalities}

The missing modality problem has motivated several research directions. \textit{Alignment-based approaches} train models to produce consistent representations across modality combinations. Ma et al.~\cite{smil} use meta-learning, Chen et al.~\cite{m3care} employ mutual information maximization, and Wang et al.~\cite{shaspec} learn shared and specific features through distribution alignment. While effective, these methods implicitly enforce similarity without explicitly identifying what information is missing.

\textit{Imputation-based methods} synthesize missing features through generative models, often producing noisy pseudo-features in high-dimensional spaces. \textit{Joint distribution learning} methods like Zhou et al.~\cite{ldcvae} learn joint posteriors through conditional VAEs, enabling conditional generation but without disentangling modality-specific from shared information. While disentanglement has been studied in multimodal learning~\cite{daunhawer2023identifiability,lee2021private,fedi}, these approaches focus on representation interpretability rather than deterministic recovery. Similarly, disentanglement of domain-invariant from domain-specific features has proven effective in medical image segmentation for domain generalization~\cite{dimix}, validating the broader principle that separating shared from specific components yields robust representations. In parallel, regularization-based methods like VICReg~\cite{bardes2021vicreg} learn decorrelated features through variance-covariance constraints, and disentangled VAEs~\cite{betavae} separate independent latent factors through reconstruction objectives—ideas that inspire our orthogonality constraints and decomposition design. However, these methods are not designed for supervised cross-modal decomposition guided by task-specific objectives. Our approach enforces algebraic invertibility within a low-rank subspace, enabling deterministic recovery of shared components, and employs latent diffusion for generation quality.

\subsection{Diffusion Models in Medical Imaging}

Diffusion models have achieved state-of-the-art results in image synthesis~\cite{yejee}. In medical imaging, they have been applied to image-to-image translation~\cite{syndiff}, MRI reconstruction~\cite{chung2022diffusion}, and cross-modal synthesis~\cite{koch2024cross}. Recent work explores diffusion for incomplete medical data~\cite{luo2023bayesian}, though in raw data space rather than learned representations. We are the first to employ latent diffusion for generating modality-specific components in learned embedding space conditioned on cross-modal information.

\section{Method}
\label{sec:method}

\subsection{Problem Formulation}

We address multimodal survival prediction where each patient $i \in \{1, ..., N\}$ is characterized by $(t_i, \delta_i, P_i, G_i)$: observed time $t_i$, event indicator $\delta_i \in \{0, 1\}$, pathological WSI $P_i$, and genomic data $G_i$, represented as variable-length token sets:
\begin{equation}
P_i = \{p_{i}^1, \dots, p_{i}^{N_P}\}, \quad
G_i = \{g_{i}^1, \dots, g_{i}^{N_G}\},
\end{equation}
where each token is embedded into a common $D$-dimensional space.

In survival analysis, the hazard function models instantaneous risk at time $t$:
\begin{equation}
h(t) = \lim_{\Delta t \to 0} \frac{Pr(t \leq T < t + \Delta t | T \geq t)}{\Delta t}
\end{equation}
which relates to the survival function $S(t) = Pr(T > t)$ through:
\begin{equation}
S(t) = \exp\left(-\int_0^t h(u) du\right) = \exp(-H(t))
\end{equation}
where $H(t)$ is the cumulative hazard function.

We adopt a discrete-time model by partitioning time into $K$ intervals with boundaries $0 = \tau_0 < \tau_1 < \cdots < \tau_K = \infty$. Each observation $t_i$ maps to interval $k_i = \max\{k : \tau_k \leq t_i\}$. The model predicts discrete hazard probabilities:
\begin{equation}
h_k = P(T = k | T \geq k, x)
\end{equation}
where $x$ represents multimodal features. The discrete survival probability is:
\begin{equation}
S(k) = \prod_{j=1}^{k} (1 - h_j)
\end{equation}

The model is trained to maximize the likelihood of observed data, which leads to the following negative log-likelihood loss:
\begin{equation}
\mathcal{L}_{\text{surv}} = -\sum_{i=1}^{N} \left[ \delta_i \log h_{k_i} + (1-\delta_i) \log S(k_i) \right]
\end{equation}
This formulation naturally handles both censored and uncensored observations: for uncensored cases ($\delta_i=1$), the loss penalizes incorrect hazard predictions at the event time, while for censored cases ($\delta_i=0$), it penalizes incorrect survival probability estimates.

\subsection{Data Preparation and Feature Extraction}

Our approach integrates two complementary modalities: WSI capturing morphological patterns and genomic profiles revealing molecular alterations. For pathology, each gigapixel WSI undergoes tissue segmentation and is divided into non-overlapping patches. We employ UNI2~\cite{uni} to extract patch-level features, yielding $P = \{p_1, \ldots, p_{N_P}\}$ where each $p_k \in \mathbb{R}^{D_P}$ is projected to dimension $D$, resulting in $P \in \mathbb{R}^{N_P \times D}$. For genomic data, we integrate RNA-seq, copy number variations (CNV), and single nucleotide variations (SNV). Rather than treating genes independently~\cite{mcat, cmta}, we organize them into functional groups based on cancer biology roles~\cite{hallmarks}. Group-specific MLPs process each group to yield genomic tokens $G = \{g_1, \ldots, g_{N_G}\} \in \mathbb{R}^{N_G \times D}$.

Our framework employs a hierarchical approach: (1) intra-modal aggregation via self-attention to obtain global representations $g_P, g_G \in \mathbb{R}^D$, (2) cross-modal interaction through bidirectional attention to extract contextualized components $c_{P \leftarrow G}, c_{G \leftarrow P} \in \mathbb{R}^D$, and (3) algebraic decomposition to separate modality-specific components $u_P, u_G \in \mathbb{R}^D$.

For pathological tokens $P$, we employ TransMIL~\cite{transmil} with a learnable class token to obtain $g_P \in \mathbb{R}^D$. For genomic tokens, we use multi-head self-attention to obtain $g_G \in \mathbb{R}^D$. To capture complementary information, we perform bidirectional cross-attention:
\begin{equation}
\textbf{CA}(Q, K, V) = \text{softmax}\left(\frac{Q W_Q (K W_K)^T}{\sqrt{d_k}}\right) V W_V
\end{equation}
where $W_Q$, $W_K$, $W_V$ are learnable projection matrices. Pathology features attend to genomics yielding $c_{P \leftarrow G}$, while genomics attends to pathology yielding $c_{G \leftarrow P}$.

Additional self-attention layers with modality-specific class tokens $[\text{CLS}_{u_P}]$ and $[\text{CLS}_{u_G}]$ extract the modality-specific components $u_P, u_G \in \mathbb{R}^D$. These learned class tokens later serve as structural priors for diffusion-based generation.

Without loss of generality, we present our method for two modalities (pathology and genomics), though the framework extends to $N$ modalities through pairwise cross-attention and shared subspace construction.

\subsection{Multimodal Algebraic Decomposition}

Representations from different modalities naturally reside in distinct semantic spaces, making direct algebraic operations problematic. While large-scale pretraining can implicitly align these spaces~\cite{word2vec}, medical datasets are typically orders of magnitude smaller. To enable algebraic operations that reflect the set-theoretic relationships between modality information, we introduce a low-rank shared subspace where cross-modal information can be meaningfully combined.

We construct this subspace through a learnable low-rank projection matrix:
\begin{equation}
P_{\cap} = B_{\cap}B_{\cap}^T, \quad B_{\cap} \in \mathbb{R}^{D \times r}
\end{equation}
where $r \ll D$ is the rank parameter. This projection operator satisfies $P_{\cap}^2 = P_{\cap}$ (idempotent), ensuring that projected vectors remain in the subspace. We decompose cross-attended and modality-specific components:
\begin{equation}
\begin{aligned}
\hat{c}_{P \leftarrow G} &= P_{\cap} c_{P \leftarrow G}, \quad \hat{c}_{G \leftarrow P} = P_{\cap} c_{G \leftarrow P} \\
\hat{u}_P &= (I - P_{\cap}) u_P, \quad \hat{u}_G = (I - P_{\cap}) u_G
\end{aligned}
\end{equation}
where the hat symbol indicates projection onto the corresponding subspace defined by $P_{\cap}$.

We enforce that the global representation of each modality can be algebraically decomposed into specific and shared components:
\begin{equation}
g_P = \hat{u}_P + \hat{c}_{G \leftarrow P}, \quad
g_G = \hat{u}_G + \hat{c}_{P \leftarrow G}
\end{equation}
To make this decomposition meaningful, we impose three constraints:

\begin{enumerate}
    \item \textit{Shared consistency}: $\hat{c}_{P \leftarrow G} = \hat{c}_{G \leftarrow P}$
    \item \textit{Inter-modal orthogonality}: $\hat{u}_P \perp \hat{u}_G$
    \item \textit{Intra-modal orthogonality}: $\hat{u}_m \perp \hat{c}_m$ for $m \in \{P, G\}$
\end{enumerate}

For survival prediction with complete data, we concatenate the modality-specific components from each modality with the shared component and feed them to a prediction head implemented as a lightweight MLP:
\begin{equation}
h_{\text{score}} = \phi([\hat{u}_P; \hat{c}; \hat{u}_G])
\end{equation}
where $\hat{c} = \hat{c}_{P \leftarrow G} = \hat{c}_{G \leftarrow P}$ under the shared consistency constraint.

\subsection{Progressive Training Strategy}

We employ a two-stage progressive training strategy to ensure stable convergence.

\subsubsection*{Stage 1 $-$ Task-Relevant Feature Learning} 
In the first stage, we focus on learning task-relevant signals from each modality with Gaussian noise injection:
\begin{equation}
\mathcal{L}_{\text{warm}} =
\mathcal{L}_{\text{surv}}(\phi([g_P; \epsilon_P])) +
\mathcal{L}_{\text{surv}}(\phi([g_G; \epsilon_G])),
\end{equation}
where $\epsilon_P, \epsilon_G \sim \mathcal{N}(0, \sigma^2 I)$ and $[\cdot;\cdot]$ denotes concatenation. The noise injection serves as a regularization mechanism that prevents overfitting to spurious correlations and ensures each encoder learns features with sufficient signal-to-noise ratio to dominate over random perturbations, thereby capturing truly informative patterns for survival prediction.

\subsubsection*{Stage 2 $-$ Algebraic Decomposition Learning}
Once encoders have learned meaningful features, we introduce the decomposition framework. The decomposition loss enforces consistency:
\begin{equation}
\begin{aligned}
\mathcal{L}_{\text{decomp}} = &\|g_P - (\hat{u}_P + \hat{c}_{G \leftarrow P})\|_2^2 + \|g_G - (\hat{u}_G + \hat{c}_{P \leftarrow G})\|_2^2
\end{aligned}
\end{equation}

The shared consistency loss ensures convergence of shared components:
\begin{equation}
\mathcal{L}_{\text{shared}} = \|\hat{c}_{P \leftarrow G} - \hat{c}_{G \leftarrow P}\|_2^2
\end{equation}

The orthogonality loss ensures independence between components:
\begin{align}
\mathcal{L}_{\text{orth}} = &|\langle \hat{u}_P, \hat{u}_G \rangle| + |\langle \hat{u}_P, \hat{c}_{P \leftarrow G} \rangle| + |\langle \hat{u}_G, \hat{c}_{G \leftarrow P} \rangle|
\end{align}
where $\langle \cdot, \cdot \rangle$ denotes the cosine similarity. The overall objective is:
\begin{equation}
\mathcal{L}_{\text{main}} = \mathcal{L}_{\text{surv}} + \lambda_{\text{dec}} \mathcal{L}_{\text{decomp}} + \lambda_{\text{sh}} \mathcal{L}_{\text{shared}} + \lambda_{\text{orth}} \mathcal{L}_{\text{orth}}
\end{equation}

\begin{figure}[t]
  \centering
  \includegraphics[width=1.0\linewidth]{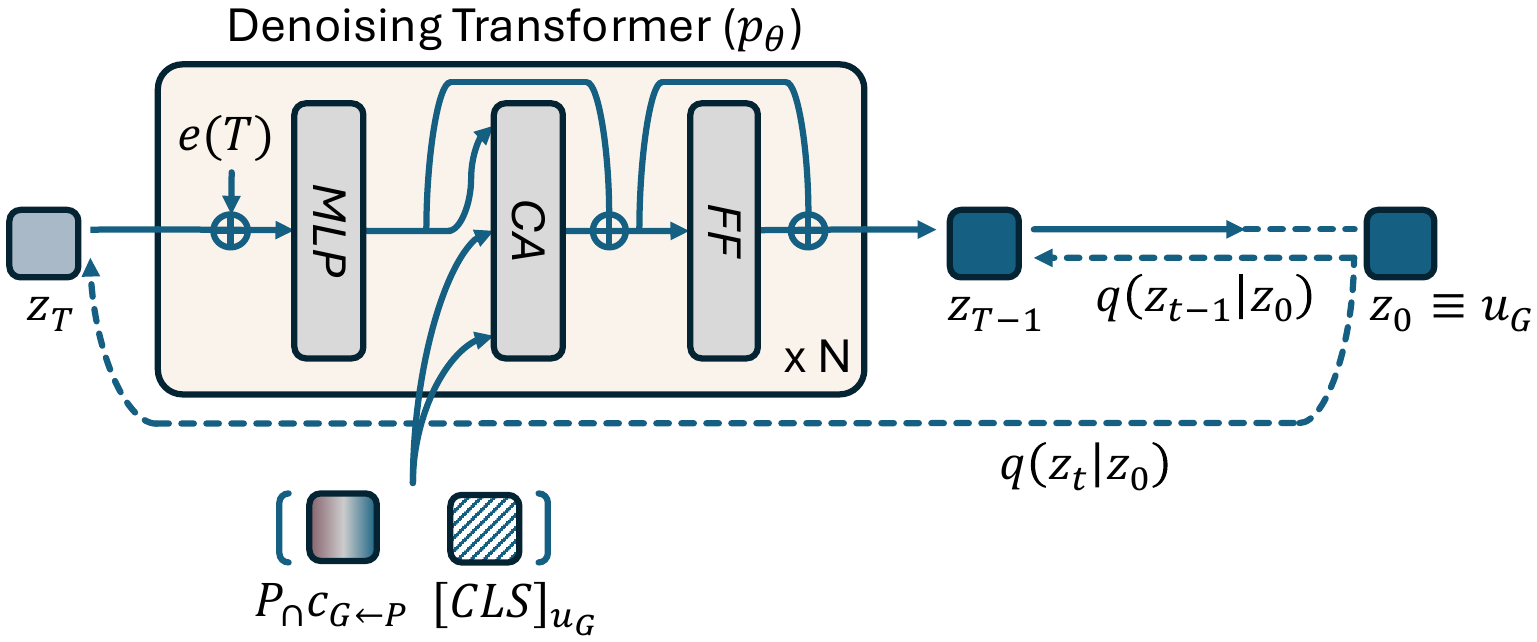}
  \caption{Conditional Diffusion Model}
  \label{fig:LDM}
\end{figure}

\subsection{Latent Diffusion Model for Missing Modality}
After the main network converges, we freeze all parameters and train conditional latent diffusion models to reconstruct missing modality-specific components. When genomics is missing, the LDM generates $\hat{u}_G$ conditioned on the shared component $\hat{c}_{G \leftarrow P}$ (derived from pathology) and the modality-specific class token $[\text{CLS}_{u_G}]$. The shared component provides cross-modal context, while the class token encodes a structural prior.

The forward diffusion process gradually adds Gaussian noise:
\begin{equation}
q(z_t | z_{t-1}) = \mathcal{N}(z_t; \sqrt{1-\beta_t}z_{t-1}, \beta_t I)
\end{equation}
where $\beta_t$ is the noise schedule and $z_0$ represents the target component to be generated.

The reverse process learns to denoise conditioned on both the shared component and class token:
\begin{equation}
p_\theta(z_{t-1} | z_t, cond) = \mathcal{N}(z_{t-1}; \mu_\theta(z_t, t, cond), \Sigma_\theta(z_t, t, cond))
\end{equation}
where $cond = [\hat{c}; [\text{CLS}_u]]$ concatenates the conditioning information, and $\theta$ denotes the learnable parameters.

The training objective minimizes:
\begin{equation}
\mathcal{L}_{\text{LDM}} = \mathbb{E}_{t,z_0,\epsilon} \left[ \|\epsilon - \epsilon_\theta(z_t, t, cond)\|_2^2 \right]
\end{equation}
where $\epsilon \sim \mathcal{N}(0, I)$ and $\epsilon_\theta$ is a transformer-based denoising network as shown in Fig.~\ref{fig:LDM}.

\subsection{Inference with Missing Modality}
\label{sec:missing}

During inference with missing modalities, we leverage the learned algebraic relationships and trained diffusion models. Given only pathology data $P$, we extract $g_P$ and $\hat{u}_P$ via the trained encoders. Using $g_P = \hat{u}_P + \hat{c}_{G \leftarrow P}$, we obtain:
\begin{equation}
\tilde{c} = g_P - \hat{u}_P = g_P - (I - P_{\cap}) u_P,
\end{equation}
which coincides with the shared component in the ideal decomposition. Empirically, high decomposition fidelity ($\cos(g, \hat{u} + \hat{c}) = 0.75$--$0.94$ across datasets; see Sec.~\ref{sec:analysis}) confirms minimal recovery error. The missing $\hat{u}_G$ is then reconstructed via DDIM sampling:
\begin{equation}
\begin{aligned}
z_{t-1} &=
\sqrt{\alpha_{t-1}}\, \tilde{z}_0(z_t, t, [\tilde{c}; [\text{CLS}_{u_G}]]) \\
&+ \sqrt{1 - \alpha_{t-1}}\, \epsilon_\theta(z_t, t, [\tilde{c}; [\text{CLS}_{u_G}]]),
\end{aligned}
\end{equation}
where $z_T \sim \mathcal{N}(0, I)$, $\alpha_t = \prod_{i=1}^{t} (1 - \beta_i)$, and $\tilde{z}_0(\cdot)$ is the DDIM estimate. Sampling proceeds from $t = T$ to $t = 0$ to obtain $\tilde{u}_G = z_0$.

The survival risk score is computed using available and reconstructed components:
\begin{equation}
h_{\text{score}} = \phi([\hat{u}_P; \tilde{c}; \tilde{u}_G]).
\end{equation}

The process is symmetric for missing pathology: given only $G$, we derive $\tilde{c} = g_G - \hat{u}_G$ and reconstruct $\tilde{u}_P$ via a pathology-specific diffusion model conditioned on $[\tilde{c}; [\text{CLS}_{u_P}]]$.

\section{Experiments}
\label{sec:experiments}

\begin{table*}[t]
\centering
\small
\setlength{\tabcolsep}{3pt}
\renewcommand{\arraystretch}{1.1}
\caption{C-index comparison across five TCGA datasets with ($\bullet$) and without (-) modalities. $\textbf{P}$ and $\textbf{G}$ denote pathological and genomic modalities, respectively. $\dagger$ indicates it requires identical modalities for training and inference. The bests are in \textbf{bold} and the second-bests are \underline{underlined}.}
\label{tab:comparison}
\begin{tabularx}{\textwidth}{@{}Ycc*5{Y}Y@{}}
\toprule
\multirow{2}{*}{\textbf{Method}} & \multirow{2}{*}{\textbf{P}}& \multirow{2}{*}{\textbf{G}}& \textbf{BLCA}& \textbf{BRCA}& \textbf{GBMLGG}& \textbf{LUAD}& \textbf{UCEC}& \multirow{2}{*}{\textbf{Overall}}\\
& & & (N=347) & (N=899) & (N=546) & (N=415) & (N=444) &\\
\hline\hline
 ABMIL$\dagger$ & $\bullet$ & - & 0.611$\pm$0.032& 0.670$\pm$0.063& 0.834$\pm$0.031& 0.633$\pm$0.054& 0.751$\pm$0.052& 0.700\\
 TransMIL$\dagger$ & $\bullet$ & - & 0.650$\pm$0.050& 0.635$\pm$0.055& 0.828$\pm$0.015& 0.611$\pm$0.032& 0.754$\pm$0.085& 0.696\\
\cmidrule(r){1-3}\cmidrule(lr){4-8}\cmidrule(l){9-9}
 MLP$\dagger$ & - & $\bullet$ & 0.678$\pm$0.045& 0.640$\pm$0.061& 0.851$\pm$0.022& 0.662$\pm$0.031& 0.612$\pm$0.046& 0.689\\
 SNN$\dagger$ & - & $\bullet$ & 0.662$\pm$0.045& 0.632$\pm$0.090& 0.854$\pm$0.022& 0.661$\pm$0.032& 0.669$\pm$0.027& 0.700\\
\cmidrule(r){1-3}\cmidrule(lr){4-8}\cmidrule(l){9-9}
 SurvPath$\dagger$ & $\bullet$ & $\bullet$ & 0.657$\pm$0.021& \underline{0.681$\pm$0.019}& \underline{0.857$\pm$0.012}& 0.652$\pm$0.033& 0.720$\pm$0.037& 0.713\\
 CMTA$\dagger$ & $\bullet$ & $\bullet$ & \underline{0.691$\pm$0.050}& 0.648$\pm$0.031& 0.857$\pm$0.022& \underline{0.667$\pm$0.049}& \underline{0.755$\pm$0.065}& \underline{0.724}\\
 MUST & $\bullet$ & $\bullet$ & \textbf{0.703$\pm$0.027}& \textbf{0.690$\pm$0.056}& \textbf{0.864$\pm$0.021}& \textbf{0.686$\pm$0.026}& \textbf{0.768$\pm$0.047}& \textbf{0.742}\\
\bottomrule
\end{tabularx}
\end{table*}

\begin{table*}[t]
\centering
\small
\setlength{\tabcolsep}{3pt}
\renewcommand{\arraystretch}{1.1}
\caption{C-index comparison across five TCGA datasets with observed ($\bullet$) and missing ($\circ$) modalities. The bests are in \textbf{bold} and the second-bests are \underline{underlined}.}
\label{tab:comparison_missing}
\begin{tabularx}{\textwidth}{@{}Ycc*5{Y}Y@{}}
\toprule
\multirow{2}{*}{\textbf{Method}} & \multirow{2}{*}{\textbf{P}}& \multirow{2}{*}{\textbf{G}}& \textbf{BLCA}& \textbf{BRCA}& \textbf{GBMLGG}& \textbf{LUAD}& \textbf{UCEC}& \multirow{2}{*}{\textbf{Overall}}\\
& & & (N=347) & (N=899) & (N=546) & (N=415) & (N=444) &\\
\hline\hline
 SMIL& $\bullet$ & $\circ$ & 0.604$\pm$0.083& 0.585$\pm$0.107& 0.690$\pm$0.070& 0.593$\pm$0.075& 0.650$\pm$0.119& 0.624\\
 M$^3$Care& $\bullet$ & $\circ$ & 0.646$\pm$0.031& 0.649$\pm$0.068& 0.789$\pm$0.051& \textbf{0.673$\pm$0.044}& 0.690$\pm$0.049& 0.689\\
 ShaSpec & $\bullet$ & $\circ$ & 0.613$\pm$0.062& 0.620$\pm$0.074& 0.816$\pm$0.024& 0.606$\pm$0.041& \underline{0.727$\pm$0.028}& 0.676\\
 LD-CVAE & $\bullet$ & $\circ$ & \underline{0.651$\pm$0.038}& \underline{0.649$\pm$0.067}& \underline{0.831$\pm$0.024}& 0.629$\pm$0.051& 0.726$\pm$0.043& \underline{0.697}\\
 MUST & $\bullet$ & $\circ$ & \textbf{0.673$\pm$0.027}& \textbf{0.651$\pm$0.059}& \textbf{0.864$\pm$0.024}& \underline{0.637$\pm$0.051}& \textbf{0.755$\pm$0.056}& \textbf{0.716}\\
\cmidrule(l){1-3}\cmidrule(lr){4-8}\cmidrule(lr){9-9}
 SMIL& $\circ$ & $\bullet$ & \underline{0.654$\pm$0.023}& 0.611$\pm$0.110& \underline{0.839$\pm$0.014}& 0.626$\pm$0.053& \underline{0.703$\pm$0.095}& \underline{0.687}\\
 M$^3$Care& $\circ$ & $\bullet$ & 0.644$\pm$0.051& \underline{0.643$\pm$0.061}& 0.797$\pm$0.033& \underline{0.653$\pm$0.053}& 0.677$\pm$0.068& 0.683\\
 ShaSpec & $\circ$ & $\bullet$ & 0.636$\pm$0.039& 0.629$\pm$0.078& 0.823$\pm$0.032& 0.610$\pm$0.029& 0.682$\pm$0.059& 0.676\\
 LD-CVAE & $\circ$ & $\bullet$ & -& -& -& -& -& -\\
 MUST & $\circ$ & $\bullet$ & \textbf{0.702$\pm$0.020}& \textbf{0.692$\pm$0.063}& \textbf{0.865$\pm$0.018}& \textbf{0.690$\pm$0.032}& \textbf{0.748$\pm$0.047}& \textbf{0.739}\\
\cmidrule(l){1-3}\cmidrule(lr){4-8}\cmidrule(lr){9-9} SMIL& $\bullet$ & $\bullet$ & 0.675$\pm$0.022& 0.641$\pm$0.072& 0.844$\pm$0.014& \textbf{0.695$\pm$0.020}& 0.740$\pm$0.059& 0.719\\
 M$^3$Care& $\bullet$ & $\bullet$ & 0.664$\pm$0.035& 0.656$\pm$0.054& 0.839$\pm$0.024& 0.683$\pm$0.036& 0.709$\pm$0.043& 0.710\\
 ShaSpec & $\bullet$ & $\bullet$ & 0.652$\pm$0.022& 0.647$\pm$0.076& \underline{0.851$\pm$0.026}& 0.648$\pm$0.027& \underline{0.747$\pm$0.046}& 0.709\\
 LD-CVAE & $\bullet$ & $\bullet$ & \underline{0.687$\pm$0.029}& \underline{0.686$\pm$0.028}& 0.845$\pm$0.021& 0.665$\pm$0.037& 0.734$\pm$0.037& \underline{0.723}\\
 MUST & $\bullet$ & $\bullet$ & \textbf{0.703$\pm$0.027}& \textbf{0.690$\pm$0.056}& \textbf{0.864$\pm$0.021}& \underline{0.686$\pm$0.026}& \textbf{0.768$\pm$0.047}& \textbf{0.742}\\
\bottomrule
\end{tabularx}
\end{table*}

\subsection{Datasets and Implementation Details}

We evaluate MUST on five TCGA cancer datasets: BLCA (N=347), BRCA (N=899), GBMLGG (N=546), LUAD (N=415), and UCEC (N=444). We use paired WSIs at 20$\times$ magnification and molecular data (RNA-Seq, CNV, SNV) to evaluate overall survival. Following previous work~\cite{mcat, cmta}, genes are organized into six functional groups based on cancer hallmarks~\cite{hallmarks}: Tumor suppression, Oncogenesis, Protein kinases, Cellular differentiation, Transcription factors, and Cytokines and growth factors.

Both pathology and genomic features are projected to $D=256$. The shared subspace rank is $r=64$. Stage 1 trains for 30 epochs with survival loss and noise injection ($\sigma=0.1$); Stage 2 trains for 30 epochs with the full objective ($\lambda_{\text{dec}}=1.0$, $\lambda_{\text{sh}}=1.0$, $\lambda_{\text{orth}}=0.5$). The LDM uses a 4-layer transformer denoising network on 256-dim features, trained for 1M steps. At inference, 50-step DDIM sampling generates each missing component. To account for sampling stochasticity, we generate 5 samples per instance and average the resulting features before feeding to the prediction head. We use Adam optimizer with learning rate $2 \times 10^{-4}$, weight decay $1 \times 10^{-5}$, batch size 1 with 32 gradient accumulation steps~\cite{clam}.

We perform 5-fold cross-validation and report C-index with standard deviation. For missing modality evaluation, all models are trained on complete paired data with missingness artificially induced at inference, thereby isolating biological signals from clinical artifacts. Statistical significance is assessed via Log-rank tests on Kaplan-Meier curves (Fig.~\ref{fig:KMcurve}). On an NVIDIA A6000 GPU, MUST requires $\leq$70ms per patient with complete data and 879ms (50 DDIM steps $\times$ 5 samples) with 1.8GB memory when a modality is missing, which remains clinically acceptable for survival prediction used in treatment planning rather than real-time decision-making.

\subsection{Comparison with State-of-the-Art Methods}

We compare MUST against unimodal and multimodal baselines. Unimodal methods include ABMIL~\cite{abmil} and TransMIL~\cite{transmil} for pathology, and MLP and SNN~\cite{snn} for genomics. Multimodal methods include SurvPath~\cite{survpath} and CMTA~\cite{cmta} which require complete data. For missing modality handling, we compare against SMIL~\cite{smil}, M$^3$Care~\cite{m3care}, ShaSpec~\cite{shaspec}, and LD-CVAE~\cite{ldcvae}.

Tab.~\ref{tab:comparison} shows MUST achieves 0.742 overall C-index, outperforming CMTA (0.724) by 2.5\%. Despite CMTA employing cross-modal attention similar to MUST, the lack of explicit decomposition limits its ability to fully capture complementary information—cross-attention alone is insufficient without the algebraic framework that prevents modality collapse. MUST shows consistent gains across cancer types (0.8-2.8\%), outperforming unimodal baselines by 6.0-6.6\%, validating algebraic decomposition as a geometric inductive bias for multimodal integration. Notably, even with complete data, the explicit separation of shared and specific components acts as a strong structural regularizer: by projecting shared information into a low-rank subspace and enforcing orthogonality between components, MUST prevents redundant feature encoding across modalities and encourages each encoder to capture maximally informative and complementary signals.

Tab.~\ref{tab:comparison_missing} reveals fundamental differences in handling incomplete data. When genomics is missing, MUST achieves 0.716 (3.5\% degradation), outperforming LD-CVAE (0.697). While ShaSpec conceptually separates shared and specific information, its reliance on head distillation matches posterior distributions without explicit algebraic structure, leading to larger degradation (4.7\%). Our key distinction lies in algebraic invertibility—the shared component is deterministically derivable from any available modality (Eq.~20), a principled recovery path absent in ShaSpec. Alignment-based methods like SMIL suffer catastrophic drops (13.2\%), confirming that implicit similarity enforcement cannot compensate for missing information.

When pathology is missing, MUST achieves 0.739 (0.4\% degradation) while maintaining bidirectional symmetry. Critically, LD-CVAE cannot handle this scenario due to its unidirectional architecture, and CMTA offers no mechanism for missing modality reconstruction. This highlights the essential advantage of MUST's algebraic framework that precisely identifies what is missing and enables deterministic reconstruction.

Comparing missing scenarios, MUST shows asymmetric degradation: 0.4\% for missing pathology versus 3.5\% for missing genomics. Notably, several datasets show improvements under missing pathology: BRCA (0.692 vs.\ 0.690), GBMLGG (0.865 vs.\ 0.864), and LUAD (0.690 vs.\ 0.686). We attribute this to different feature characteristics: when generating $\hat{u}_P$ from 10K-200K noisy patches, the LDM's denoising produces regularized features filtering high-frequency noise, while genomic $\hat{u}_G$ from six structured groups already contains precise representations. The smoothing inherent in LDM generation, while beneficial for noisy pathology, may attenuate functionally important variations in well-structured genomic features. Importantly, this predictability does not imply redundancy: through Eqs.~9 and 15, specific components reside in mutually orthogonal subspaces, making them linearly unreachable from one another (Sec.~\ref{sec:analysis}).

\subsection{Patient Stratification Analysis}

\begin{figure}[t]
  \centering
  \includegraphics[width=1.0\linewidth]{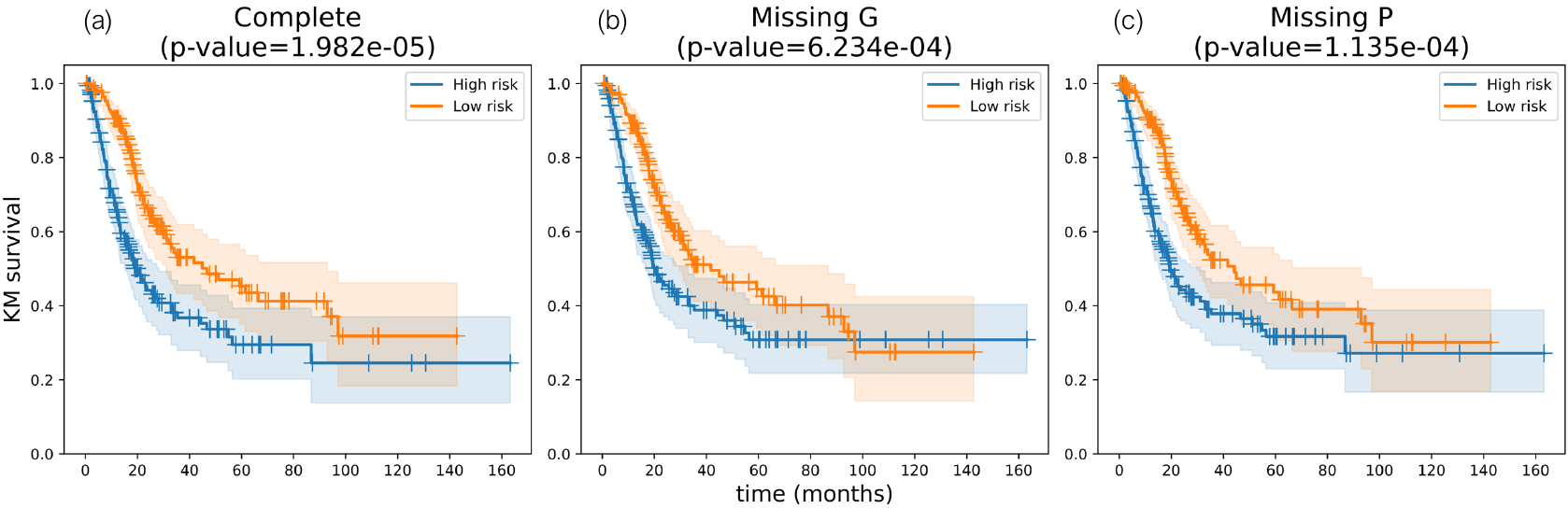}
  \caption{Kaplan-Meier survival curves on BLCA dataset comparing high-risk and low-risk groups: (a) complete data, (b) missing $G$, (c) missing $P$. Shaded regions represent 95\% confidence intervals.}
  \label{fig:KMcurve}
\end{figure}

Fig.~\ref{fig:KMcurve} shows Kaplan-Meier curves for the BLCA dataset, stratifying patients into high- and low-risk groups at the median predicted risk score. Under complete data (Fig.~\ref{fig:KMcurve}a), MUST achieves clear separation with p=1.982e-05, demonstrating that the predicted risk scores capture meaningful prognostic differences. Critically, this stratification is preserved under missing genomics (Fig.~\ref{fig:KMcurve}b, p=6.234e-04) and missing pathology (Fig.~\ref{fig:KMcurve}c, p=1.135e-04), demonstrating that reconstructed components retain clinically meaningful prognostic information for risk group identification. The maintained statistical significance across all missing scenarios confirms that MUST's two-stage reconstruction—deterministic shared recovery followed by diffusion-based specific generation—preserves the relative risk ordering essential for clinical decision-making, even when an entire modality is absent. Similar results hold across all five datasets (see supplementary), where MUST consistently maintains significant stratification while baseline methods such as SMIL and ShaSpec lose statistical significance on multiple datasets.

\subsection{Ablation Studies and Analysis}
\label{sec:analysis}


\begin{table}[t]
\centering
\caption{Ablation studies on progressive training and LDM conditioning strategies. $\circ$ indicates missing modality.}
\label{tab:ablation}
\small
\begin{minipage}[t]{0.43\columnwidth}
\centering
\subcaption{Progressive training}
\label{tab:warmup}
\resizebox{\linewidth}{!}{
\begin{tabular}{l|cc}
\toprule
\multirow{2}{*}{Dataset} & \multicolumn{2}{c}{Warm-up} \\
& w/o & w/ \\
\midrule
BLCA & 0.684 & \textbf{0.703} \\
BRCA & 0.647 & \textbf{0.690} \\
GBMLGG & 0.862 & \textbf{0.864} \\
LUAD & \textbf{0.690} & 0.686 \\
UCEC & 0.733 & \textbf{0.768} \\
\bottomrule
\end{tabular}
}
\end{minipage}
\hfill
\begin{minipage}[t]{0.48\columnwidth}
\centering
\subcaption{LDM conditioning}
\label{tab:ldm_cond}
\resizebox{\linewidth}{!}{
\begin{tabular}{cc|cc}
\toprule
\multicolumn{2}{c|}{$[\hat{c}]$} & \multicolumn{2}{c}{$[\hat{c};[\text{CLS}]]$} \\
$\circ$G & $\circ$P & $\circ$G & $\circ$P \\
\midrule
0.665 & 0.695 & \textbf{0.673} & \textbf{0.702} \\
\textbf{0.656} & 0.684 & 0.651 & \textbf{0.692} \\
0.856 & 0.860 & \textbf{0.864} & \textbf{0.865} \\
0.636 & 0.682 & \textbf{0.637} & \textbf{0.690} \\
0.746 & 0.740 & \textbf{0.755} & \textbf{0.748} \\
\bottomrule
\end{tabular}
}
\end{minipage}
\end{table}

\begin{figure*}[t]
  \centering
   \includegraphics[width=1.0\linewidth]{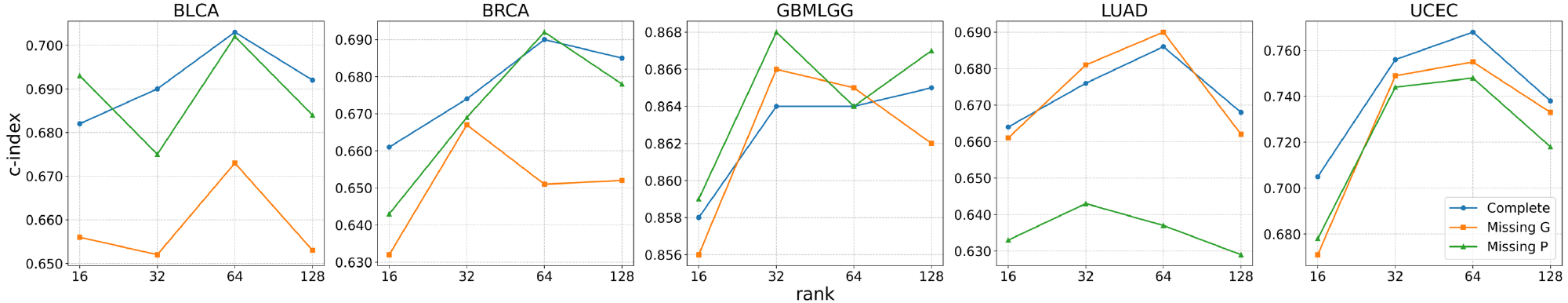}
   \caption{Ablation on rank $r$ of the shared subspace $P_\cap$.}
   \label{fig:rank}
\end{figure*}

\begin{figure}[t]
  \centering
  \includegraphics[width=1.0\linewidth]{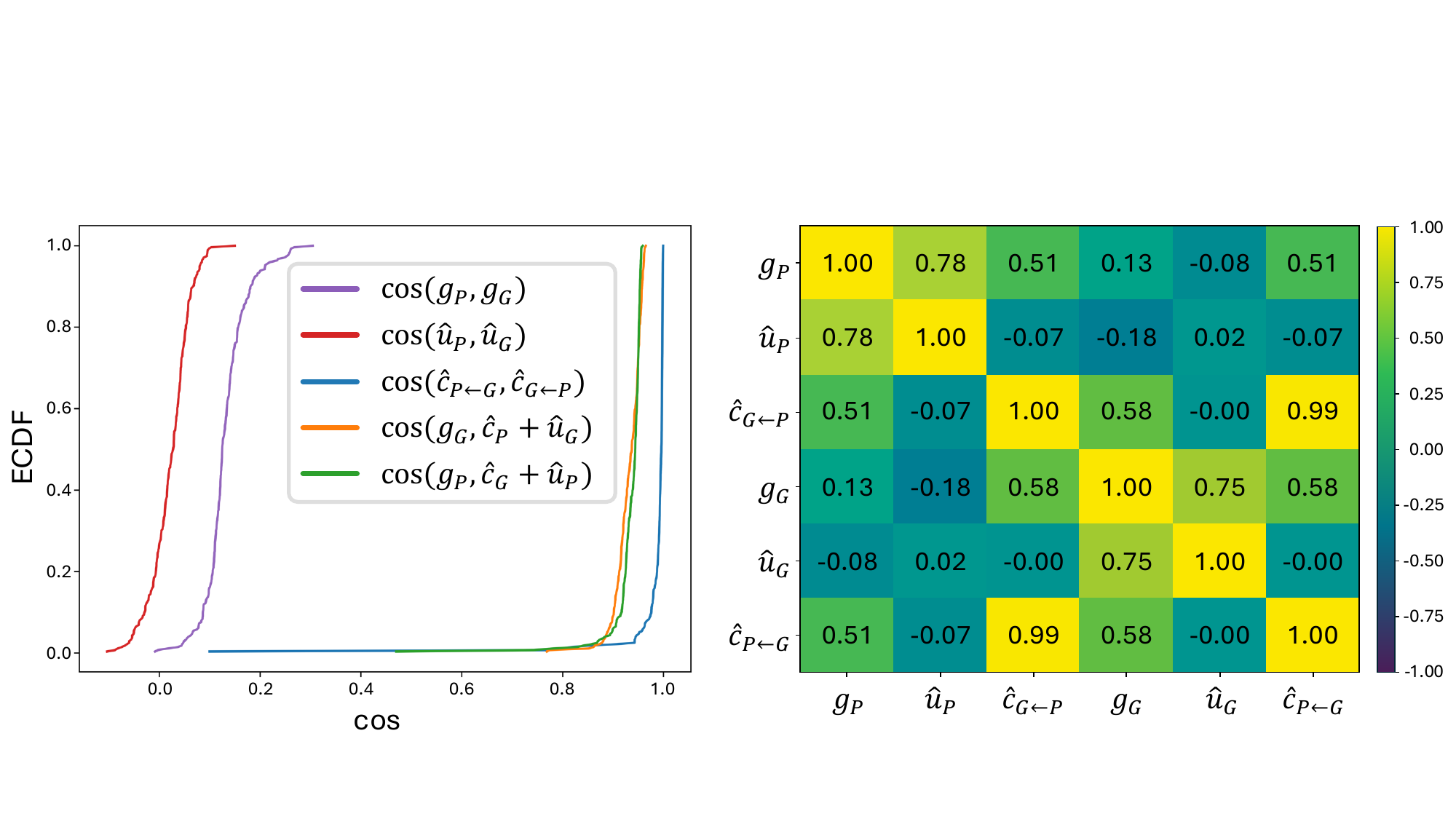}
  \caption{ECDF curve and cosine similarity map on UCEC dataset.}
  \label{fig:cossim}
\end{figure}

\textbf{Progressive Training with Warm-up.} Tab.~\ref{tab:warmup} compares models with and without the Stage 1 warm-up phase. Without warm-up, directly training the full decomposition from random initialization leads to unstable learning and 1-3\% degradation. The warm-up allows encoders to first learn task-relevant global representations before introducing algebraic constraints, preventing component collapse.

\noindent\textbf{Impact of Low-Rank Shared Subspace.} Fig.~\ref{fig:rank} examines how rank $r$ of the shared subspace projection $P_{\cap} = B_{\cap}B_{\cap}^T$ affects performance across $r \in \{16, 32, 64, 128\}$. Performance peaks at $r=64$, achieving optimal balance between expressiveness and regularization. Lower ranks are too constrained for shared cross-modal information, while $r=128$ approaches the full $D=256$ space, losing decomposition benefits. This demonstrates that enforcing a low-rank bottleneck is crucial for learning meaningful algebraic relationships.

\noindent\textbf{Conditioning Strategy for Latent Diffusion.} Tab.~\ref{tab:ldm_cond} evaluates the effect of including the structural class token $[\text{CLS}_u]$ as conditioning signal. Conditioning on $[\hat{c}; [\text{CLS}_u]]$ consistently outperforms conditioning only on $[\hat{c}]$, with improvements of 0.8-1.5\%. The shared component captures most cross-modal information, while the class token provides additional structural guidance for precise feature distributions.

\noindent\textbf{Effectiveness of Algebraic Decomposition.} Fig.~\ref{fig:cossim} validates the learned algebraic constraints on the UCEC dataset. The ECDF for $\cos(\hat{u}_P, \hat{u}_G)$ (red) concentrates near zero ($-0.08$ on average), confirming that modality-specific components are nearly orthogonal as intended by $\mathcal{L}_{\text{orth}}$. The curve for $\cos(\hat{c}_{P \leftarrow G}, \hat{c}_{G \leftarrow P})$ (green) concentrates near 1.0 ($0.99$ on average), verifying that $\mathcal{L}_{\text{shared}}$ successfully aligns shared components across modalities. The cross-modality similarity $\cos(g_P, g_G) = 0.13$ confirms that the two modalities capture complementary rather than redundant information. The algebraic relations $g_P \approx \hat{u}_P + \hat{c}_{G \leftarrow P}$ and $g_G \approx \hat{u}_G + \hat{c}_{P \leftarrow G}$ are confirmed by the cosine similarity map: $\cos(g_P, \hat{u}_P + \hat{c}_{G \leftarrow P}) = 0.78$ and $\cos(g_G, \hat{u}_G + \hat{c}_{P \leftarrow G}) = 0.75$ on UCEC, with 0.80--0.94 across other datasets (see supplementary). These high reconstruction fidelities validate that the subtraction-based recovery in Eq.~20 operates with minimal error, providing the foundation for reliable missing modality inference. Additional analysis including hyperparameter sensitivity and inference cost is in the supplementary.

\section{Conclusion}

We presented MUST for robust multimodal survival prediction under missing modalities. By explicitly decomposing representations into modality-specific and cross-modal contextualized components through algebraic constraints in a low-rank subspace, MUST enables precise identification and reconstruction of missing information via deterministic projection and LDM-based generation. The key innovation is algebraic invertibility—shared components are deterministically derivable from any available modality, isolating stochastic generation to modality-specific residuals. Experiments on five TCGA datasets show state-of-the-art complete-data performance (C-index: 0.742) with robust missing-modality predictions (0.716 and 0.739 for missing genomics and pathology), confirming MUST's potential for clinical deployment where modality availability varies.

\noindent\textbf{Limitations and Future Work.} Our framework prioritizes paired training for strict algebraic constraints. Notably, MUST's shared-specific decomposition structure naturally suggests a pathway for handling training-time missingness, and preliminary experiments show promising adaptability even with 50\% unpaired data (see supplementary). Fully addressing this setting remains future work.

\section*{Acknowledgements}
This work was supported by the National Research Foundation of Korea (NRF) grant funded by the Korea government (MSIT) (No. RS-2025-16070382, RS-2025-02215070, RS-2025-02217919), Artificial Intelligence Graduate School Program at Yonsei University (RS-2020-II201361), the Korea Institute of Science and Technology (KIST) Institutional Program under Grant 26E0170.

{
    \small
    \bibliographystyle{ieeenat_fullname}
    \bibliography{main}

\begin{thebibliography}{43}
\providecommand{\natexlab}[1]{#1}
\providecommand{\url}[1]{\texttt{#1}}
\expandafter\ifx\csname urlstyle\endcsname\relax
  \providecommand{\doi}[1]{doi: #1}\else
  \providecommand{\doi}{doi: \begingroup \urlstyle{rm}\Url}\fi

\bibitem[Acosta et~al.(2022)Acosta, Falcone, Rajpurkar, and Topol]{acosta2022multimodal}
Juli{\'a}n~N Acosta, Guido~J Falcone, Pranav Rajpurkar, and Eric~J Topol.
\newblock Multimodal biomedical ai.
\newblock \emph{Nature medicine}, 28\penalty0 (9):\penalty0 1773--1784, 2022.

\bibitem[Baltru{\v{s}}aitis et~al.(2018)Baltru{\v{s}}aitis, Ahuja, and Morency]{baltrusaitis2018multimodal}
Tadas Baltru{\v{s}}aitis, Chaitanya Ahuja, and Louis-Philippe Morency.
\newblock Multimodal machine learning: A survey and taxonomy.
\newblock \emph{IEEE transactions on pattern analysis and machine intelligence}, 41\penalty0 (2):\penalty0 423--443, 2018.

\bibitem[Bardes et~al.(2021)Bardes, Ponce, and LeCun]{bardes2021vicreg}
Adrien Bardes, Jean Ponce, and Yann LeCun.
\newblock Vicreg: Variance-invariance-covariance regularization for self-supervised learning.
\newblock \emph{arXiv preprint arXiv:2105.04906}, 2021.

\bibitem[Campanella et~al.(2019)Campanella, Hanna, Geneslaw, Miraflor, Werneck Krauss~Silva, Busam, Brogi, Reuter, Klimstra, and Fuchs]{campanella2019clinical}
Gabriele Campanella, Matthew~G Hanna, Luke Geneslaw, Allen Miraflor, Vitor Werneck Krauss~Silva, Klaus~J Busam, Edi Brogi, Victor~E Reuter, David~S Klimstra, and Thomas~J Fuchs.
\newblock Clinical-grade computational pathology using weakly supervised deep learning on whole slide images.
\newblock \emph{Nature medicine}, 25\penalty0 (8):\penalty0 1301--1309, 2019.

\bibitem[Cheerla and Gevaert(2019)]{cheerla2019deep}
Anika Cheerla and Olivier Gevaert.
\newblock Deep learning with multimodal representation for pancancer prognosis prediction.
\newblock \emph{Bioinformatics}, 35\penalty0 (14):\penalty0 i446--i454, 2019.

\bibitem[Chen et~al.(2020)Chen, Lu, Wang, Williamson, Rodig, Lindeman, and Mahmood]{chen2020pathomic}
Richard~J Chen, Ming~Y Lu, Jingwen Wang, Drew~FK Williamson, Scott~J Rodig, Neal~I Lindeman, and Faisal Mahmood.
\newblock Pathomic fusion: an integrated framework for fusing histopathology and genomic features for cancer diagnosis and prognosis.
\newblock \emph{IEEE Transactions on Medical Imaging}, 41\penalty0 (4):\penalty0 757--770, 2020.

\bibitem[Chen et~al.(2021)Chen, Lu, Weng, Chen, Williamson, Manz, Shady, and Mahmood]{mcat}
Richard~J Chen, Ming~Y Lu, Wei-Hung Weng, Tiffany~Y Chen, Drew~FK Williamson, Trevor Manz, Maha Shady, and Faisal Mahmood.
\newblock Multimodal co-attention transformer for survival prediction in gigapixel whole slide images.
\newblock In \emph{Proceedings of the IEEE/CVF international conference on computer vision}, pages 4015--4025, 2021.

\bibitem[Chen et~al.(2022)Chen, Lu, Williamson, Chen, Lipkova, Noor, Shaban, Shady, Williams, Joo, et~al.]{chen2022pan}
Richard~J Chen, Ming~Y Lu, Drew~FK Williamson, Tiffany~Y Chen, Jana Lipkova, Zahra Noor, Muhammad Shaban, Maha Shady, Mane Williams, Bumjin Joo, et~al.
\newblock Pan-cancer integrative histology-genomic analysis via multimodal deep learning.
\newblock \emph{Cancer cell}, 40\penalty0 (8):\penalty0 865--878, 2022.

\bibitem[Chen et~al.(2024)Chen, Ding, Lu, Williamson, Jaume, Song, Chen, Zhang, Shao, Shaban, et~al.]{uni}
Richard~J Chen, Tong Ding, Ming~Y Lu, Drew~FK Williamson, Guillaume Jaume, Andrew~H Song, Bowen Chen, Andrew Zhang, Daniel Shao, Muhammad Shaban, et~al.
\newblock Towards a general-purpose foundation model for computational pathology.
\newblock \emph{Nature medicine}, 30\penalty0 (3):\penalty0 850--862, 2024.

\bibitem[Chung et~al.(2022)Chung, Kim, Mccann, Klasky, and Ye]{chung2022diffusion}
Hyungjin Chung, Jeongsol Kim, Michael~T Mccann, Marc~L Klasky, and Jong~Chul Ye.
\newblock Diffusion posterior sampling for general noisy inverse problems.
\newblock \emph{arXiv preprint arXiv:2209.14687}, 2022.

\bibitem[Daunhawer et~al.(2023)Daunhawer, Bizeul, Palumbo, Marx, and Vogt]{daunhawer2023identifiability}
Imant Daunhawer, Alice Bizeul, Emanuele Palumbo, Alexander Marx, and Julia~E Vogt.
\newblock Identifiability results for multimodal contrastive learning.
\newblock In \emph{International Conference on Learning Representations}, 2023.

\bibitem[Dorent et~al.(2019)Dorent, Joutard, Modat, Ourselin, and Vercauteren]{dorent2019hetero}
Reuben Dorent, Samuel Joutard, Marc Modat, S{\'e}bastien Ourselin, and Tom Vercauteren.
\newblock Hetero-modal variational encoder-decoder for joint modality completion and segmentation.
\newblock In \emph{International Conference on Medical Image Computing and Computer-Assisted Intervention}, pages 74--82. Springer, 2019.

\bibitem[Girdhar et~al.(2023)Girdhar, El-Nouby, Liu, Singh, Alwala, Joulin, and Misra]{imagebind}
Rohit Girdhar, Alaaeldin El-Nouby, Zhuang Liu, Mannat Singh, Kalyan~Vasudev Alwala, Armand Joulin, and Ishan Misra.
\newblock Imagebind: One embedding space to bind them all.
\newblock In \emph{Proceedings of the IEEE/CVF conference on computer vision and pattern recognition}, pages 15180--15190, 2023.

\bibitem[Hanahan and Weinberg(2011)]{hallmarks}
Douglas Hanahan and Robert~A Weinberg.
\newblock Hallmarks of cancer: the next generation.
\newblock \emph{cell}, 144\penalty0 (5):\penalty0 646--674, 2011.

\bibitem[Higgins et~al.(2017)Higgins, Matthey, Pal, Burgess, Glorot, Botvinick, Mohamed, and Lerchner]{betavae}
Irina Higgins, Loic Matthey, Arka Pal, Christopher Burgess, Xavier Glorot, Matthew Botvinick, Shakir Mohamed, and Alexander Lerchner.
\newblock beta-vae: Learning basic visual concepts with a constrained variational framework.
\newblock In \emph{International conference on learning representations}, 2017.

\bibitem[Ho et~al.(2020)Ho, Jain, and Abbeel]{ddpm}
Jonathan Ho, Ajay Jain, and Pieter Abbeel.
\newblock Denoising diffusion probabilistic models.
\newblock \emph{Advances in neural information processing systems}, 33:\penalty0 6840--6851, 2020.

\bibitem[Ilse et~al.(2018)Ilse, Tomczak, and Welling]{abmil}
Maximilian Ilse, Jakub Tomczak, and Max Welling.
\newblock Attention-based deep multiple instance learning.
\newblock In \emph{International conference on machine learning}, pages 2127--2136. PMLR, 2018.

\bibitem[Jaume et~al.(2024)Jaume, Vaidya, Chen, Williamson, Liang, and Mahmood]{survpath}
Guillaume Jaume, Anurag Vaidya, Richard~J Chen, Drew~FK Williamson, Paul~Pu Liang, and Faisal Mahmood.
\newblock Modeling dense multimodal interactions between biological pathways and histology for survival prediction.
\newblock In \emph{Proceedings of the IEEE/CVF Conference on Computer Vision and Pattern Recognition}, pages 11579--11590, 2024.

\bibitem[Kim et~al.(2023)Kim, Shin, and Hwang]{dimix}
Hyeongyu Kim, Yejee Shin, and Dosik Hwang.
\newblock Dimix: Disentangle-and-mix based domain generalizable medical image segmentation.
\newblock In \emph{International Conference on Medical Image Computing and Computer-Assisted Intervention}, pages 242--251. Springer, 2023.

\bibitem[Klambauer et~al.(2017)Klambauer, Unterthiner, Mayr, and Hochreiter]{snn}
G{\"u}nter Klambauer, Thomas Unterthiner, Andreas Mayr, and Sepp Hochreiter.
\newblock Self-normalizing neural networks.
\newblock \emph{Advances in neural information processing systems}, 30, 2017.

\bibitem[Koch et~al.(2024)Koch, Aydin, Hilbert, Rieger, Tanioka, Ishida, and Frey]{koch2024cross}
Alexander Koch, Orhun~Utku Aydin, Adam Hilbert, Jana Rieger, Satoru Tanioka, Fujimaro Ishida, and Dietmar Frey.
\newblock Cross-modality image synthesis from tof-mra to cta using diffusion-based models.
\newblock \emph{arXiv preprint arXiv:2409.10089}, 2024.

\bibitem[Lee et~al.()Lee, Son, and Hwang]{fedi}
Jeongryong Lee, Geonhui Son, and Dosik Hwang.
\newblock Fedi: Feature disentanglement for self-supervised learning.
\newblock \emph{Available at SSRN 5184015}.

\bibitem[Lee and Pavlovic(2021)]{lee2021private}
Mihee Lee and Vladimir Pavlovic.
\newblock Private-shared disentangled multimodal {VAE} for learning of latent representations.
\newblock In \emph{Proceedings of the IEEE/CVF Conference on Computer Vision and Pattern Recognition}, pages 1692--1700, 2021.

\bibitem[Li et~al.(2022)Li, Zhu, Yao, and Huang]{li2022hierarchical}
Chunyuan Li, Xinliang Zhu, Jiawen Yao, and Junzhou Huang.
\newblock Hierarchical transformer for survival prediction using multimodality whole slide images and genomics.
\newblock In \emph{2022 26th international conference on pattern recognition (ICPR)}, pages 4256--4262. IEEE, 2022.

\bibitem[Lu et~al.(2021)Lu, Williamson, Chen, Chen, Barbieri, and Mahmood]{clam}
Ming~Y Lu, Drew~FK Williamson, Tiffany~Y Chen, Richard~J Chen, Matteo Barbieri, and Faisal Mahmood.
\newblock Data-efficient and weakly supervised computational pathology on whole-slide images.
\newblock \emph{Nature biomedical engineering}, 5\penalty0 (6):\penalty0 555--570, 2021.

\bibitem[Luo et~al.(2023)Luo, Blumenthal, Heide, and Uecker]{luo2023bayesian}
Guanxiong Luo, Moritz Blumenthal, Martin Heide, and Martin Uecker.
\newblock Bayesian mri reconstruction with joint uncertainty estimation using diffusion models.
\newblock \emph{Magnetic Resonance in Medicine}, 90\penalty0 (1):\penalty0 295--311, 2023.

\bibitem[Ma et~al.(2021)Ma, Ren, Zhao, Tulyakov, Wu, and Peng]{smil}
Mengmeng Ma, Jian Ren, Long Zhao, Sergey Tulyakov, Cathy Wu, and Xi Peng.
\newblock Smil: Multimodal learning with severely missing modality.
\newblock In \emph{Proceedings of the AAAI conference on artificial intelligence}, pages 2302--2310, 2021.

\bibitem[Mikolov et~al.(2013)Mikolov, Chen, Corrado, and Dean]{word2vec}
Tomas Mikolov, Kai Chen, Greg Corrado, and Jeffrey Dean.
\newblock Efficient estimation of word representations in vector space.
\newblock \emph{arXiv preprint arXiv:1301.3781}, 2013.

\bibitem[{\"O}zbey et~al.(2023){\"O}zbey, Dalmaz, Dar, Bedel, {\"O}zturk, G{\"u}ng{\"o}r, and Cukur]{syndiff}
Muzaffer {\"O}zbey, Onat Dalmaz, Salman~UH Dar, Hasan~A Bedel, {\c{S}}aban {\"O}zturk, Alper G{\"u}ng{\"o}r, and Tolga Cukur.
\newblock Unsupervised medical image translation with adversarial diffusion models.
\newblock \emph{IEEE Transactions on Medical Imaging}, 42\penalty0 (12):\penalty0 3524--3539, 2023.

\bibitem[Radford et~al.(2021)Radford, Kim, Hallacy, Ramesh, Goh, Agarwal, Sastry, Askell, Mishkin, Clark, et~al.]{radford2021learning}
Alec Radford, Jong~Wook Kim, Chris Hallacy, Aditya Ramesh, Gabriel Goh, Sandhini Agarwal, Girish Sastry, Amanda Askell, Pamela Mishkin, Jack Clark, et~al.
\newblock Learning transferable visual models from natural language supervision.
\newblock In \emph{International conference on machine learning}, pages 8748--8763. PmLR, 2021.

\bibitem[Rombach et~al.(2022)Rombach, Blattmann, Lorenz, Esser, and Ommer]{ldm}
Robin Rombach, Andreas Blattmann, Dominik Lorenz, Patrick Esser, and Bj{\"o}rn Ommer.
\newblock High-resolution image synthesis with latent diffusion models.
\newblock In \emph{Proceedings of the IEEE/CVF conference on computer vision and pattern recognition}, pages 10684--10695, 2022.

\bibitem[Shao et~al.(2021)Shao, Bian, Chen, Wang, Zhang, Ji, et~al.]{transmil}
Zhuchen Shao, Hao Bian, Yang Chen, Yifeng Wang, Jian Zhang, Xiangyang Ji, et~al.
\newblock Transmil: Transformer based correlated multiple instance learning for whole slide image classification.
\newblock \emph{Advances in neural information processing systems}, 34:\penalty0 2136--2147, 2021.

\bibitem[Sharma and Hamarneh(2019)]{sharma2019missing}
Anmol Sharma and Ghassan Hamarneh.
\newblock Missing mri pulse sequence synthesis using multi-modal generative adversarial network.
\newblock \emph{IEEE transactions on medical imaging}, 39\penalty0 (4):\penalty0 1170--1183, 2019.

\bibitem[Shin et~al.(2025)Shin, Lee, Jang, Son, Kim, and Hwang]{yejee}
Yejee Shin, Yeeun Lee, Hanbyol Jang, Geonhui Son, Hyeongyu Kim, and Dosik Hwang.
\newblock Anatomical consistency and adaptive prior-informed transformation for multi-contrast mr image synthesis via diffusion model.
\newblock In \emph{Proceedings of the Computer Vision and Pattern Recognition Conference}, pages 30918--30927, 2025.

\bibitem[Song et~al.(2020)Song, Meng, and Ermon]{ddim}
Jiaming Song, Chenlin Meng, and Stefano Ermon.
\newblock Denoising diffusion implicit models.
\newblock \emph{arXiv preprint arXiv:2010.02502}, 2020.

\bibitem[Vale-Silva and Rohr(2021)]{vale2021long}
Lu{\'\i}s~A Vale-Silva and Karl Rohr.
\newblock Long-term cancer survival prediction using multimodal deep learning.
\newblock \emph{Scientific Reports}, 11\penalty0 (1):\penalty0 13505, 2021.

\bibitem[Wang et~al.(2023{\natexlab{a}})Wang, Chen, Ma, Avery, Hull, and Carneiro]{shaspec}
Hu Wang, Yuanhong Chen, Congbo Ma, Jodie Avery, Louise Hull, and Gustavo Carneiro.
\newblock Multi-modal learning with missing modality via shared-specific feature modelling.
\newblock In \emph{Proceedings of the IEEE/CVF conference on computer vision and pattern recognition}, pages 15878--15887, 2023{\natexlab{a}}.

\bibitem[Wang et~al.(2023{\natexlab{b}})Wang, Xiao, Bi, Li, and Gao]{wang2023mcf}
Yongchao Wang, Bin Xiao, Xiuli Bi, Weisheng Li, and Xinbo Gao.
\newblock Mcf: Mutual correction framework for semi-supervised medical image segmentation.
\newblock In \emph{Proceedings of the IEEE/CVF conference on computer vision and pattern recognition}, pages 15651--15660, 2023{\natexlab{b}}.

\bibitem[Xu and Chen(2023)]{motcat}
Yingxue Xu and Hao Chen.
\newblock Multimodal optimal transport-based co-attention transformer with global structure consistency for survival prediction.
\newblock In \emph{Proceedings of the IEEE/CVF international conference on computer vision}, pages 21241--21251, 2023.

\bibitem[Zhang et~al.(2022)Zhang, Chu, Ma, Zhu, Wang, Wang, and Zhao]{m3care}
Chaohe Zhang, Xu Chu, Liantao Ma, Yinghao Zhu, Yasha Wang, Jiangtao Wang, and Junfeng Zhao.
\newblock M3care: Learning with missing modalities in multimodal healthcare data.
\newblock In \emph{Proceedings of the 28th ACM SIGKDD conference on knowledge discovery and data mining}, pages 2418--2428, 2022.

\bibitem[Zhang et~al.(2023)Zhang, Xu, Usuyama, Xu, Bagga, Tinn, Preston, Rao, Wei, Valluri, et~al.]{biomedclip}
Sheng Zhang, Yanbo Xu, Naoto Usuyama, Hanwen Xu, Jaspreet Bagga, Robert Tinn, Sam Preston, Rajesh Rao, Mu Wei, Naveen Valluri, et~al.
\newblock Biomedclip: a multimodal biomedical foundation model pretrained from fifteen million scientific image-text pairs.
\newblock \emph{arXiv preprint arXiv:2303.00915}, 2023.

\bibitem[Zhou and Chen(2023)]{cmta}
Fengtao Zhou and Hao Chen.
\newblock Cross-modal translation and alignment for survival analysis.
\newblock In \emph{Proceedings of the IEEE/CVF International Conference on Computer Vision}, pages 21485--21494, 2023.

\bibitem[Zhou et~al.(2025)Zhou, Tang, Zuo, Wan, Zhang, and Shao]{ldcvae}
Junjie Zhou, Jiao Tang, Yingli Zuo, Peng Wan, Daoqiang Zhang, and Wei Shao.
\newblock Robust multimodal survival prediction with conditional latent differentiation variational autoencoder.
\newblock In \emph{Proceedings of the Computer Vision and Pattern Recognition Conference}, pages 10384--10393, 2025.

\end{thebibliography}
}

\clearpage
\maketitlesupplementary

The Supplementary Material provides detailed implementation settings, training configurations, and additional experimental results.

\section{Implementation Details}

\subsection{Datasets}
We evaluate our method on five cancer datasets from The Cancer Genome Atlas (TCGA): Bladder Urothelial Carcinoma (BLCA, $N=347$), Breast Invasive Carcinoma (BRCA, $N=899$), Glioblastoma and Low-Grade Glioma (GBMLGG, $N=546$), Lung Adenocarcinoma (LUAD, $N=415$), and Uterine Corpus Endometrial Carcinoma (UCEC, $N=444$). For genomic data, we integrate three molecular data types: RNA-seq expression profiles, copy number variations (CNV), and single nucleotide variations (SNV). Samples with missing values in any of these genomic modalities are excluded from our study to ensure fair comparison across methods. Following the cancer hallmarks framework~\cite{hallmarks}, we organize genes into six biologically meaningful functional groups: Tumor suppression, Oncogenesis, Protein kinases, Cellular differentiation, Transcription factors, and Cytokines and growth factors.

\subsection{Survival Time Discretization}
For the discrete-time survival model described in Sec.~3.1 of the main paper, we partition the continuous time axis into $K=4$ intervals. The interval boundaries are determined by ensuring that uncensored samples are distributed uniformly across bins, with censored samples subsequently assigned to the appropriate bins based on their observation times. This binning strategy balances computational efficiency with sufficient temporal resolution for survival prediction.

\subsection{Feature Extraction}
For pathological data, we employ the UNI2-h histopathology foundation model~\cite{uni} to extract patch-level features from whole slide images at 20$\times$ magnification. Each patch is encoded into a 1536-dimensional feature vector, which is then projected to the common embedding dimension of $D=256$ through a linear layer. For genomic data, we do not use pretrained encoders. Instead, each of the six functional gene groups is processed by a group-specific multi-layer perceptron (MLP) that learns to encode the corresponding gene expression patterns into 256-dimensional embeddings. These genomic token encoders are trained end-to-end as part of the survival prediction task.

\subsection{Architecture Details}
As described in Sec.~3 (Method), our framework processes pathological and genomic tokens differently. For pathological tokens $P = \{p_1, \ldots, p_{N_P}\}$, we apply TransMIL~\cite{transmil} with learnable class tokens to obtain global representations $g_P$, modality-specific components $u_P$, and cross-attended components $c_{P \leftarrow G}$. For genomic tokens $G = \{g_1, \ldots, g_{N_G}\}$, we prepend modality-specific class tokens and apply multi-head self-attention layers to extract $g_G$, $u_G$, and $c_{G \leftarrow P}$. The low-rank shared subspace projection $P_{\cap} = B_{\cap}B_{\cap}^T$ uses rank $r=64$ as determined by ablation studies in Sec.~4 (Experiments).

\section{Training Configurations}

\subsection{Main Training}
Due to the variable number of patches in whole slide images (ranging from 10K to 200K patches per slide), we set the batch size to 1 for all experiments. To ensure training stability, we employ gradient accumulation with 32 accumulation steps, effectively simulating a batch size of 32. Our two-stage progressive training strategy trains Stage 1 (warm-up) for 30 epochs using only the survival loss with Gaussian noise injection ($\sigma=0.1$), followed by Stage 2 (main) for 30 epochs with the full objective including decomposition, shared consistency, and orthogonality losses. We use the AdamW optimizer with weight decay $1 \times 10^{-5}$ for both stages. In Stage 1, we set the learning rate to $1 \times 10^{-3}$, and in Stage 2, we reduce it to $2 \times 10^{-4}$ to facilitate stable learning of the algebraic constraints. Loss weights are set to $\lambda_{\text{dec}}=1.0$, $\lambda_{\text{sh}}=1.0$, and $\lambda_{\text{orth}}=0.5$ as specified in the main paper.

\subsection{Latent Diffusion Model Training}
After the main network converges, we freeze all encoder and decomposition parameters and train conditional latent diffusion models~\cite{ldm} to reconstruct missing modality-specific components. The denoising network is implemented as a 4-layer transformer architecture operating on 256-dimensional feature embeddings. We train separate diffusion models for pathology-specific and genomic-specific component generation, each trained for 1 million steps using the AdamW optimizer with learning rate $1 \times 10^{-4}$ and batch size 32. During inference, we perform 50-step DDIM~\cite{ddim} sampling to generate missing components. To account for the stochastic nature of the sampling process, we generate 5 samples per test instance and average the resulting features before feeding them to the prediction head, as described in Sec.~4 (Experiments).

\subsection{Comparison Methods}
For fair comparison, we implement all baseline methods following their original papers while adapting them to our experimental setup. All experiments are conducted with batch size 1 due to the variable number of patches in whole slide images. SurvPath~\cite{survpath} is trained for 30 epochs using the Adam optimizer with learning rate $1 \times 10^{-3}$. CMTA~\cite{cmta} is trained for 30 epochs using SGD with learning rate $1 \times 10^{-3}$. For methods handling missing modalities, SMIL~\cite{smil} is trained for 50 epochs with Adam optimizer (learning rate $1 \times 10^{-3}$), M$^3$Care~\cite{m3care} for 30 epochs with Adam (learning rate $1 \times 10^{-3}$), ShaSpec~\cite{shaspec} for 30 epochs with Adam (learning rate $1 \times 10^{-3}$), and LD-CVAE~\cite{ldcvae} for 30 epochs with Adam (learning rate $2 \times 10^{-4}$). All methods are trained until the training C-index fully converges to ensure optimal performance.

\subsection{Evaluation Protocol}
We perform 5-fold cross-validation for all experiments and report the concordance index (C-index) along with standard deviations across folds. For each fold, we split the data at the patient level to ensure no data leakage. Statistical significance of risk group stratification is assessed using the log-rank test on Kaplan-Meier survival curves with the median predicted risk score as the threshold for defining high-risk and low-risk groups.

\section{Additional Experimental Results}

\subsection{Hyperparameter Sensitivity Analysis}

\begin{figure}[t]
  \centering
   \includegraphics[width=1.0\linewidth]{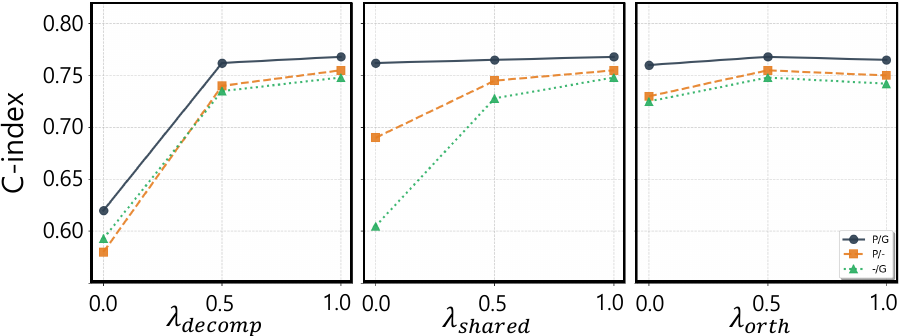}
   \caption{Hyperparameter sensitivity analysis on the UCEC dataset. Each loss weight ($\lambda_{\text{decomp}}$, $\lambda_{\text{shared}}$, $\lambda_{\text{orth}}$) is varied independently while keeping others at their default values. P/G denotes complete data, P/- denotes missing genomics, and -/G denotes missing pathology.}
   \label{fig:lambda_supp}
\end{figure}

Fig.~\ref{fig:lambda_supp} evaluates the sensitivity of MUST to the three loss weights $\lambda_{\text{dec}}$, $\lambda_{\text{sh}}$, and $\lambda_{\text{orth}}$ on the UCEC dataset. Each weight is varied independently while keeping others at their default values ($\lambda_{\text{dec}}=1.0$, $\lambda_{\text{sh}}=1.0$, $\lambda_{\text{orth}}=0.5$).

$\mathcal{L}_{\text{decomp}}$ has the most pronounced effect: setting $\lambda_{\text{dec}}=0$ causes severe degradation across all scenarios, as the algebraic decomposition structure breaks down entirely. With non-zero values, performance stabilizes rapidly, confirming that the decomposition constraint is essential but not sensitive to precise tuning. $\mathcal{L}_{\text{shared}}$ primarily affects missing modality scenarios: removing it ($\lambda_{\text{sh}}=0$) leads to substantial drops under missing genomics and missing pathology, since cross-modal recovery relies on aligned shared components. Complete-data performance remains largely unaffected, as the shared component is not directly used during complete-data inference. $\mathcal{L}_{\text{orth}}$ shows the gentlest effect: even at $\lambda_{\text{orth}}=0$, performance degrades only moderately, though with notable modality collapse where cross-modal recovery becomes less reliable. Our selected values ($\lambda_{\text{dec}}=1.0$, $\lambda_{\text{sh}}=1.0$, $\lambda_{\text{orth}}=0.5$) balance complete-data performance with missing-modality robustness across all three loss terms.

\subsection{Training-Time Missingness}

While MUST is primarily designed for paired training with complete modalities, its architectural structure—decomposing representations into shared and modality-specific components—naturally suggests a pathway for handling training-time missingness as well. Specifically, the shared consistency constraint ($\mathcal{L}_{\text{shared}}$) aligns shared components across modalities, meaning that even when one modality is absent during training, the available modality's shared component can still provide a meaningful surrogate for cross-modal recovery. This structural property motivates us to explore whether MUST can be adapted to scenarios where complete pairs are not always available.

To evaluate this, we conduct experiments on the UCEC dataset by varying the proportion of unpaired samples (20\% and 50\% missing rate). For unpaired data, we substitute zero-tensors for the missing modality-specific component $\hat{u}$ and apply only the survival loss $\mathcal{L}_{\text{surv}}$, while training the LDM with the available $\tilde{c}$ as a surrogate condition.

\begin{table}[h]
\centering
\caption{Training-time missingness evaluation on UCEC. Values show C-index without $\rightarrow$ with LDM-based inference.}
\label{tab:train_missing}
\small
\begin{tabular}{l|cc}
\toprule
Missing Rate & Missing G & Missing P \\
\midrule
0\% (baseline) & 0.755 & 0.748 \\
20\% & 0.716 $\rightarrow$ 0.735 & 0.606 $\rightarrow$ 0.724 \\
50\% & 0.713 $\rightarrow$ 0.724 & 0.601 $\rightarrow$ 0.716 \\
\bottomrule
\end{tabular}
\end{table}

Tab.~\ref{tab:train_missing} shows that MUST remains robust under training-time missingness, and the performance gain when replacing zero-tensors with LDM-sampled residuals at inference confirms that $\mathcal{L}_{\text{shared}}$ effectively aligns shared components for cross-modal recovery even with incomplete training pairs. While our framework prioritizes paired training to establish strict algebraic constraints, these results demonstrate practical adaptability to realistic clinical scenarios where complete data may not always be available during model development.

\subsection{Kaplan-Meier Survival Analysis}

To provide comprehensive validation of our method's robustness under missing modality scenarios, we present detailed Kaplan-Meier survival curves comparing MUST against representative baseline methods: ShaSpec~\cite{shaspec} and SMIL~\cite{smil}. We focus on these two methods as LD-CVAE~\cite{ldcvae}'s unidirectional architecture only supports missing genomics scenarios. We stratify patients into high-risk and low-risk groups at the median predicted risk score and assess statistical significance using the log-rank test across all five TCGA datasets.

\begin{figure*}[t]
  \centering
   \includegraphics[width=0.99\linewidth]{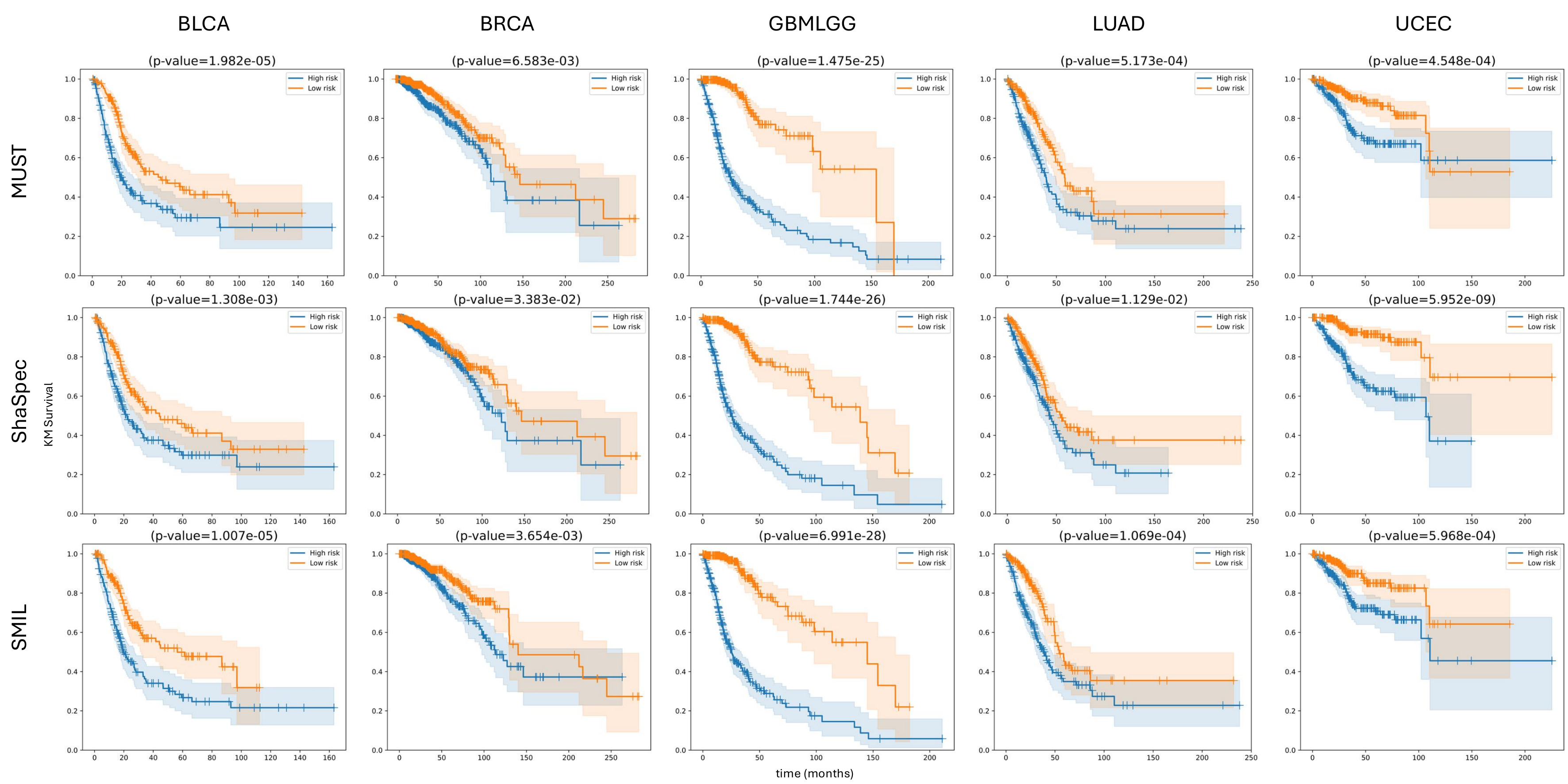}
   \caption{Kaplan-Meier survival curves on five different datasets comparing high-risk and low-risk groups across complete scenario. p-values are computed using the log-rank test. Patients are stratified into high-risk (blue) and low-risk (orange) groups based on predicted risk scores at the median threshold. Shaded regions represent 95\% confidence intervals, and censored observations are indicated by vertical tick marks on the curves.}
   \label{fig:KM_complete_supp}
\end{figure*}

When both modalities are available as shown in Fig.~\ref{fig:KM_complete_supp}, all three methods achieve statistically significant risk stratification ($p < 0.05$) across all datasets, demonstrating comparable discriminative power. The clear separation between high-risk and low-risk groups indicates that all methods effectively learn prognostic patterns when full multimodal information is available.

\begin{figure*}[h]
  \centering
   \includegraphics[width=1.0\linewidth]{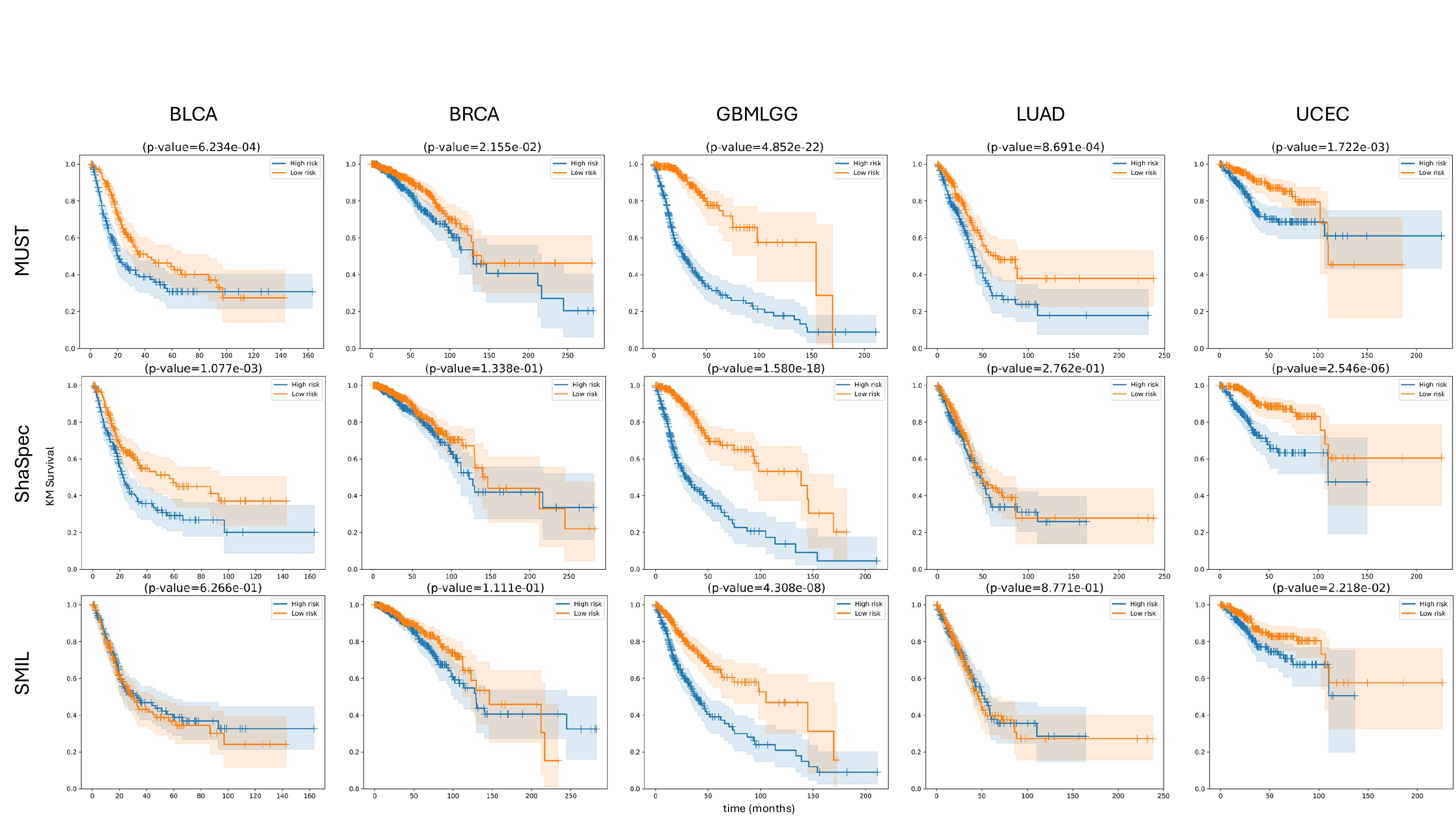}
   \caption{Kaplan-Meier survival curves on five different datasets comparing high-risk and low-risk groups across missing genomics ($G$) scenario. p-values are computed using the log-rank test. Patients are stratified into high-risk (blue) and low-risk (orange) groups based on predicted risk scores at the median threshold. Shaded regions represent 95\% confidence intervals, and censored observations are indicated by vertical tick marks on the curves.}
   \label{fig:KM_missing_G_supp}
\end{figure*}

When genomic data is unavailable as shown in Fig.~\ref{fig:KM_missing_G_supp}, substantial differences in robustness emerge. MUST maintains statistically significant stratification across all five datasets, demonstrating effective reconstruction of missing genomic-specific components. In contrast, both baselines show notable deterioration. ShaSpec loses statistical significance on BRCA ($p=1.338$e-$01$) and LUAD ($p=2.762$e-$01$), indicating weakened discriminative capability. SMIL exhibits severe vulnerability, failing to maintain significance on BLCA ($p=6.266$e-$01$), BRCA ($p=1.111$e-$01$), and LUAD ($p=8.771$e-$01$). This catastrophic degradation confirms that alignment-based implicit imputation cannot adequately compensate for missing information without explicit modeling of modality-specific structures.

\begin{figure*}[h]
  \centering
   \includegraphics[width=1.0\linewidth]{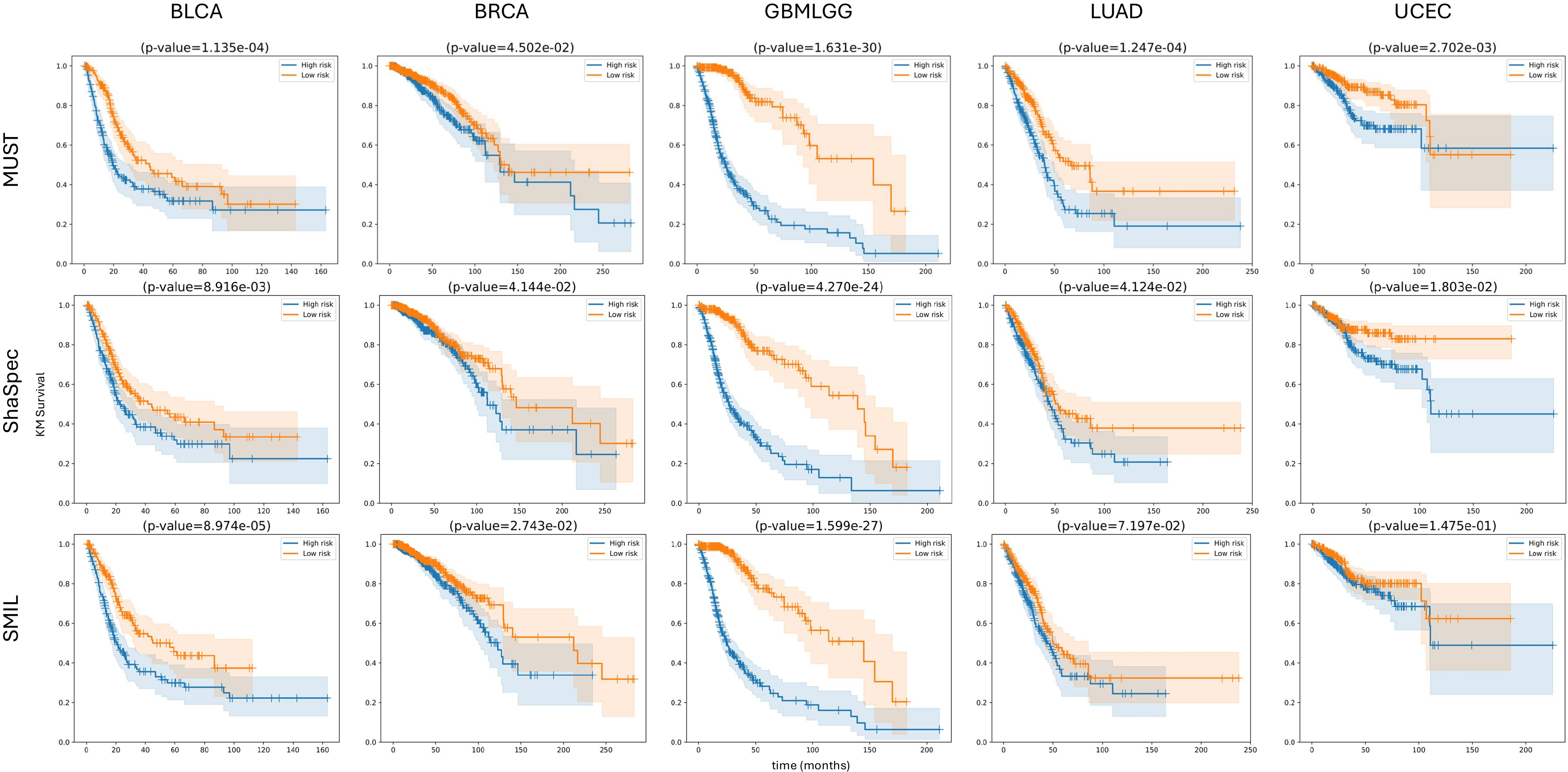}
   \caption{Kaplan-Meier survival curves on five different datasets comparing high-risk and low-risk groups across missing pathology ($P$) scenario. p-values are computed using the log-rank test. Patients are stratified into high-risk (blue) and low-risk (orange) groups based on predicted risk scores at the median threshold. Shaded regions represent 95\% confidence intervals, and censored observations are indicated by vertical tick marks on the curves.}
   \label{fig:KM_missing_P_supp}
\end{figure*}

In Fig.~\ref{fig:KM_missing_P_supp}, MUST continues to exhibit remarkable robustness when pathological images are missing, maintaining statistically significant stratification across all datasets with minimal degradation from complete data performance. This validates that our diffusion-based reconstruction of pathology-specific components preserves essential prognostic information. ShaSpec also maintains statistical significance across all datasets in this scenario, though with moderate performance degradation compared to MUST. SMIL shows intermediate vulnerability, losing significance on UCEC ($p=1.475$e-$01$) while maintaining significance on other datasets.

\subsection{Decomposition Analysis}
\begin{figure*}[h]
  \centering
   \includegraphics[width=1.0\linewidth]{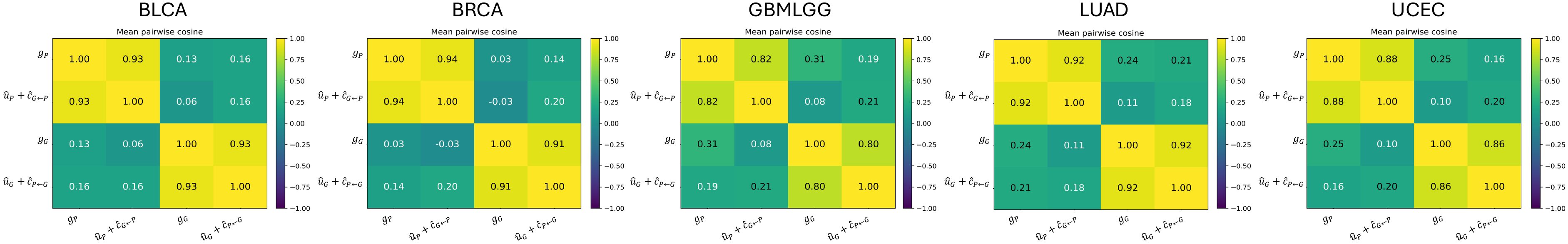}
   \caption{Cosine similarity map between $g$ and corresponding composition of $\hat{u}$ and $\hat{c}$ across all five TCGA datasets.}
   \label{fig:CosSim_supp}
\end{figure*}

Fig.~\ref{fig:CosSim_supp} presents pairwise cosine similarities between global representations and their algebraic decompositions across all five datasets. The analysis validates that our decomposition successfully reconstructs the original global representations while maintaining their semantic structure.

The key validation comes from comparing global representations with their decomposed counterparts. For pathology, the decomposition achieves high reconstruction fidelity with cosine similarities ranging from 0.82 (GBMLGG) to 0.94 (BRCA), confirming that $g_P \approx \hat{u}_P + \hat{c}_{G \leftarrow P}$ holds with high precision. For genomics, the reconstruction quality is similarly high with similarities from 0.80 (GBMLGG) to 0.93 (BLCA), validating that $g_G \approx \hat{u}_G + \hat{c}_{P \leftarrow G}$.

Cross-modality similarities between $g_P$ and $g_G$ are notably low (0.03-0.31), confirming that the two modalities capture complementary rather than redundant information. Similarly, the low similarities between $g_P$ and $\hat{u}_G + \hat{c}_{P \leftarrow G}$ (0.14-0.21) and between $g_G$ and $\hat{u}_P + \hat{c}_{G \leftarrow P}$ (0.06-0.25) further confirm that modality-specific components retain distinctive characteristics even after decomposition.

The consistently high reconstruction similarities across both modalities explain why MUST can maintain robust performance under missing modality scenarios. When a modality is missing, the decomposition framework enables deterministic reconstruction through the algebraic relation: the shared component can be derived from the available modality ($\tilde{c} = g_{\text{available}} - \hat{u}_{\text{available}}$), and the missing modality-specific component is generated via diffusion conditioning on this shared component.

\clearpage

\end{document}